\documentclass[11pt]{article}
\usepackage[pdftex]{graphicx}
\usepackage[small,bf,textfont={small}]{caption}
\usepackage{vmargin}
\usepackage{caption}
\usepackage{subcaption}
\usepackage{amsmath}
\usepackage{amssymb} 
\usepackage{fancyhdr}
\usepackage{color}
\usepackage[colorlinks=true,
            linkcolor=blue,
            urlcolor=blue,
            citecolor=blue]{hyperref}
\usepackage[numbers,sort&compress]{natbib}
\usepackage{doi}
\usepackage{booktabs}
\usepackage{float}
\usepackage{lineno}
\usepackage{epstopdf}
\usepackage{array}
\usepackage{multirow}
\usepackage{bm}
\usepackage{tikz}

\floatstyle{ruled}
\newfloat{floatbox}{thp}{lob}
\floatname{floatbox}{Algorithm}

\let\svthefootnote\thefootnote
\newcommand\freefootnote[1]{%
	\let\thefootnote\relax\footnotetext{#1}%
	\let\thefootnote\svthefootnote%
}

\begin{document}
%\linenumbers

%\title{Data Assimilation GAN (DA-GAN) applied to determine the spatial variation of COVID-19 infections through time}
\title{Data Assimilation Predictive GAN (DA-PredGAN): applied to determine the spread of COVID-19}

\author{Vinicius L S Silva$^{*,1,2}$ \and Claire E Heaney$^{1,2}$ \and Yaqi Li$^{2}$ \and Christopher C Pain$^{1,2,3}$}

\maketitle

\begin{center}
\small \noindent $^1$Applied Modelling and Computation Group, Imperial College London, UK\\
\small \noindent $^2$Department of Earth Science and Engineering, Imperial College London, UK\\
\small \noindent $^3$Data Assimilation Laboratory, Data Science Institute, Imperial College London, UK\\
\small \noindent $^*$ Corresponding author email: viluiz@gmail.com
\end{center}

\section*{Abstract}
\label{sec:abstract}
We propose the novel use of a generative adversarial network (GAN) (i)~to  make predictions in time (PredGAN) and (ii)~to assimilate measurements (DA-PredGAN). In the latter case, we take advantage of the natural adjoint-like properties of generative models and the ability to simulate forwards and backwards in time. GANs have received much attention recently, after achieving excellent results for their generation of realistic-looking images. We wish to explore how this property translates to new applications in computational modelling and to exploit the adjoint-like properties for efficient data assimilation. To predict the spread of COVID-19 in an idealised town, we apply these methods to a compartmental model in epidemiology that is able to model space and time variations. To do this, the GAN is set within a reduced-order model (ROM), which uses a low-dimensional space for the spatial distribution of the simulation states. Then the GAN learns the evolution of the low-dimensional states over time. The results show that the proposed methods can accurately predict the evolution of the high-fidelity numerical simulation, and can efficiently assimilate observed data and determine the corresponding model parameters.

\vspace{10mm}

\noindent \textbf{Keywords:} generative adversarial networks; spatio-temporal prediction; data assimilation; reduced-order model; deep learning; COVID-19.\\

%\noindent \textbf{Mathematics Subject Classification:} 68T01, 68T07, 65H10, 65M32, 92D30  

\freefootnote{Source code and data are available at \url{https://github.com/viluiz/gan}}

%\clearpage
\section{Introduction}
\label{sec:introduction}

A combination of the availability of large data sets, the advances in algorithms and the accessibility of computational power has resulted in an unparalleled surge of interest in machine learning, and subsequently significant advances have been made in many different fields. Machine learning can be seen as a process of solving practical problems by building a statistical model based on a given dataset. This building process can be broadly classified as supervised learning, when there is the presence of the outcome variable to guide the learning process, or unsupervised learning, when there are only the features and no measurements of the outcome. In the latter, the main goal is to describe how the data is organised or clustered \citep{hastie:09}. Recently, a class of machine learning methods referred to as deep learning (for either supervised or unsupervised problems) have been achieving extraordinary results, surpassing the ones obtained from previous machine learning techniques \citep{goodfellow:16,geron:19,liu:17}. Based on artificial neural networks, deep learning techniques use multiple layers to extract features or patterns progressively from the data. Examples of this can be found in pioneering work by \citet{lecun:89}, using convolutional neural networks, and by \citet{rumelhart:86}, using recurrent neural networks.

Deep generative models are one type of unsupervised learning and aim to generate samples from complex probability distributions in high-dimensional spaces \citep{goodfellow:16}. They learn the structure of the input data (which has an unknown closed form) and can be used to generate new instances that appear to have been taken from the training data. There are several types of generative models including deep belief networks (DBN) \citep{hinton:06}, variational autoencoders (VAE) \citep{kingma:13}, and the generative adversarial network (GAN) \citep{goodfellow:14}. Here, we focus our attention on the latter. A GAN comprises two networks, a generator and a discriminator. During training, the former produces samples (so-called ``fake samples") from a set of random variables, and the second network attempts to distinguish between samples drawn from the training data and the fake samples. After training, the generator can be used to produce realistic samples, and the discriminator can be used to distinguish between samples. Although originally developed within the field of image generation, more recently, applications of GANs in computational physics have emerged. These applications attempt to exploit the capabilities of GANs, and predict realistic spatially and temporally varying solutions. For example, \citet{gupta:18} tackle the problem of predicting human trajectories using a novel GAN based encoder-decoder framework. Their proposed method predicted socially plausible futures that outperformed prior works. \citet{xie:18} address the problem of super-resolution fluid flows by using a GAN to infer three-dimentional volumetric data in time. They used two discriminators, one that focuses on space while the other focuses on temporal aspects. \citet{zhong:19} proposed a conditional GAN as a surrogate model for predicting the migration of carbon dioxide plumes in heterogeneous reservoirs. Their results show high accuracy prediction in space and time when compared with a compositional reservoir simulator. \citet{cheng:20} used a GAN to make spatio-temporal predictions of a nonlinear fluid flow. They demonstrate that the results of the GAN are comparable with those from the high-fidelity numerical model. All of the previous works tackled the problem of spatio-temporal model prediction. In addition to predicting in time, for many practical applications, being able to assimilate observed data is highly desirable.

Data assimilation is an inverse problem with the aim of calibrating uncertain model parameters in order to generate results that match observed data within some tolerance \citep{tarantola:05,oliver:11b,amore:14,silva:17}. Some researchers used GANs to tackle this specific problem. \citet{mosser:19} trained a GAN to represent the prior distribution of subsurface properties and integrated it within a data assimilation framework based on adjoint capabilities.  \citet{kang:20} proposed a method where a cluster technique using principal component analysis (also known as proper orthogonal decomposition) and K-means is performed in the prior models to select realisation that match the observed data. Then a GAN is trained on these realisations in order to generate calibrated models. \citet{razak:20} used a similar approach of clustering the prior models; however, they also apply a conditional GAN to label production responses of each model. \citet{canchumuni:21} compared different deep generative networks formulations, including GANs and VAE, integrated with a Kalman filter-based method for proper data assimilation of facies models in reservoir simulations. A common characteristic of these works is that they use GANs in order to generate the model parameters. The forward simulations (spatio-temporal predictions) still need to be performed using the high-fidelity numerical simulator. In this work, we propose two contributions: the generation of spatio-temporal predictions using GANs and the assimilation of observed data using GANs. In the first contribution, an algorithm is developed so that a GAN is able to make predictions in time (PredGAN) for unseen model parameters. After the GAN has learnt the evolution of the system, an iterative process is applied to the generator in order to march forward in time. In the second contribution, the iterative process is extended in order to assimilate observed data and generate the corresponding model parameters (DA-PredGAN). No additional simulation of the high-fidelity numerical model is required during the data assimilation process adding to the efficiency of this method.

The test case chosen here is the spatio-temporal variation of a virus infection in an idealised town. Where possible, parameters of the model were chosen to be consistent with those of COVID-19. %the COVID-19 virus. 
With compartments of susceptible (S), exposed (E), infectious (I) and recovered (R), the extended SEIRS equations~\citep{quilodran:21} are used to generate the high-fidelity numerical simulations that describe the spread of infections in both space and time. Based on differential equations, multi-compartment SEIR-type models \citep{viguerie:20,viguerie:21} can be very costly to solve: there may be millions of variables every time step; and the time steps may need to be small to model the movement of people around the domain. When applied to a city, a country or the entire world, solving such models can therefore require substantial computational resources. In order to reduce the computational cost of numerical simulations, reduced-order models (ROM) are now commonly used for applications in computational physics \citep{Rozza:08,cardoso:09}. Nonetheless, they are relatively new to virus modelling \citep{quilodran:21}. A ROM is a low-dimensional representation of a high-dimensional model or discretised system, and should be accurate enough for the desired use and at least several orders of magnitude faster to solve than the high-dimensional system. Three steps are involved in their construction: (i) generation of solutions of the high-dimensional system (snapshots), (ii) compression of the snapshots to find a low-dimensional space for the approximation, and (iii) approximation of the high-dimensional system in the low-dimensional space. The low-dimensional space is often found with methods based on singular value decomposition, such as proper orthogonal decomposition (POD) \citep{strazzullo:20}, although autoencoders offer a promising alternative \citep{Phillips:20}. In the third step, the high-dimensional system is projected onto the low-dimensional space for a projection-based ROM \citep{Benner:15}, whereas for a non-intrusive ROM (NIROM), the snapshots are projected onto the low-dimensional space and the dynamics of the system in this space for unseen parameters are represented by interpolation. This can be performed by classical interpolation methods such as cubic splines~\cite{li_huang:21} or radial basis functions (RBF)~\cite{audouze:13}. The RBF approach was extended by \cite{xiao:17} who used a Smolyak grid to sample the parameter space; by \cite{kostorz:20} who interpolated values of model parameters and time levels using one parametrisation; and by \cite{alsayyari:21} who used adaptive sampling in time. Recently, neural networks have been used to perform the interpolation, and examples of this for steady-state parametrised problems can be found in \cite{hesthaven:18,dalsanto:20}, both of whom use POD and multi-layer perceptrons, and in \cite{swischuk:19}, who use POD and compare a number of different networks. Examples for time-dependent parametrised problems can be found in \cite{sugar-gabor:20}, who used feed-forward neural networks to model the viscous Burgers' equation; \cite{xu:20}, who proposed a nested trio of networks to learn spatial patterns, temporal patterns and to learn the dependence on the model parameters; \cite{maulik:21}, who combine convolutional autoencoders with recurrent neural networks for Burgers' equation and the shallow water equations; \cite{fresca:21,nikolopoulos:21}, both of whom combine an autoencoder and a feed-forward neural network; and \cite{lu:21}, who train an multi-layer perceptron with data from both high-fidelity and low-fidelity models to improve the accuracy of the model. In this paper, we set the PredGAN and DA-PredGAN algorithms within a NIROM framework, using POD for the compression step and a GAN for learning how the dynamics depend on the model parameters. However, both PredGAN and DA-PredGAN could also be used without the NIROM framework (for smaller problems).

The main novelties of this research involve the application of a new GAN approach to both spatio-temporal prediction and data assimilation. This requires an additional optimisation every time step in order to be able to use the generator within the GAN for predictions. This optimisation proves to be well suited to data assimilation problems using adjoints and gradient-based approaches. It is worth mentioning that this method is not limited to the underlying physics. It is a general framework for developing surrogate models and assimilating observed data.

This paper is structured as follows: the next section (Section \ref{sec:method}) provides the description of the proposed method for spatio-temporal prediction and data assimilation with GANs. Section \ref{sec:testcase} introduces the test case, a spatio-temporal compartmental model in epidemiology. After that, the results of the prediction and data assimilation are presented in Section \ref{sec:results}. Section \ref{sec:disc} presents some further discussions. Finally, concluding remarks are provided in Section \ref{sec:conclusion}.        

\section{Method}
\label{sec:method} 

In this section, firstly a method to make spatio-temporal predictions using a GAN is proposed. This algorithm is set within a NIROM framework in order to reduce the number of variables that the GAN has to work on, however, for problems with fewer degrees of freedom, this may not be necessary. The NIROM involves finding a low-dimensional space in which to approximate high-dimensional model snapshots of a high-fidelity numerical simulation. The GAN then learns the evolution of the numerical simulation based on the evolution of the snapshots in the low-dimensional space. Therefore, the aim of predicting in time using GANs is to be a surrogate model for the high-fidelity numerical simulation. Secondly, considering we have observed data, we can extend the forecasting using GANs to match the given data and generate the model parameters, without running any additional simulations of the high-fidelity numerical model. 

\subsection{Predicting in space and time using GANs}
\label{sec:prediction_in_time}

Proposed by \citet{goodfellow:14}, GANs are unsupervised learning algorithms capable of learning dense representations of the input data, and can be used as generative models: they are capable of generating new samples following the same distribution of the training dataset. The training is based on a game theory scenario in which the generator network $G$ must compete against an adversary. The generator network directly produces samples from a random distribution as the input (latent vector $\mathbf{z}$) and its adversary, the discriminator network $D$, attempts to distinguish between samples drawn from the training data and samples drawn from the generator. The output of the discriminator $D(\mathbf{x})$ represents the probability that a sample came from the data rather than a “fake” sample from the generator. The output of the generator $G(\mathbf{z})$ is a sample from the distribution learned from the dataset. Eqs.~\eqref{eq:Ld} and~\eqref{eq:Lg} show the loss function of the discriminator and generator used in this work, respectively,

\begin{eqnarray}
L_D & = & -\mathbb{E}_{x \sim p_{data}(x)}[\log(D(\mathbf{x}))] - \mathbb{E}_{z \sim p_{z}(z)}[\log(1- D(G(\mathbf{z})))]\,, \label{eq:Ld} \\[2mm]
L_G & = & \phantom{-}\mathbb{E}_{z \sim p_{z}(z)}[\log(1- D(G(\mathbf{z})))]\,. \label{eq:Lg}
\end{eqnarray}

In order to make predictions in space and time using a GAN, here an algorithm named Predictive GAN (PredGAN) is proposed. We train a GAN to be able to generate the following nonlinear map
\begin{equation}
G(\mathbf{z}^n) =\bm{\Phi}^n,
\label{eq:gan} 
\end{equation}
between the latent variables, $\mathbf{z}^n$ at time level~$n$, and the solution generated by the GAN $\bm{\Phi}^n$. The $\bm{\Phi}^n$ is made up of $m$ consecutive time steps of compressed variables~$\bm{\alpha}$ (compressed spatial outputs of the numerical simulation) which are proper orthogonal decomposition (POD) coefficients, but could also be latent variables from an autoencoder, and the vector $\bm{\mu}$ of parameters used within the high-fidelity model (e.g.~a material property, or other simulation input). For a GAN that has been trained with $m$ time levels, $\bm{\Phi}^n$ takes the following form
\begin{equation}\label{phi}
\bm{\Phi}^n = \left[ \begin{array}{c} 
(\bm{\alpha}^{n-m+1})^T, (\bm{\mu}^{n-m+1})^T \\[1mm]
(\bm{\alpha}^{n-m+2})^T, (\bm{\mu}^{n-m+2})^T \\[1mm]
\vdots \\[1mm]
(\bm{\alpha}^{n-1})^T, (\bm{\mu}^{n-1})^T \\[1mm]
(\bm{\alpha}^n)^T, (\bm{\mu}^n)^T \\[1mm]
\end{array}
\right],
\end{equation}
where $(\bm{\alpha}^n)^T= [\alpha_1^n, \alpha_2^n, \cdots , \alpha_{N_{\text{POD}}}^n]$ and $(\bm{\mu}^n)^T =[ \mu_1^n, \mu_2^n, \cdots, \mu_{N_\mu}^n]$. $N_{\text{POD}}$ is the number of POD coefficients, $\alpha_i^n$ represents the $i$th POD coefficient at time level $n$, $N_\mu$ is the number of model parameters, and $\mu_i^n$ represents the $i$th parameter at time level $n$.  

By design, the generator of a GAN produces realistic-looking solutions (images) from a randomly-generated set of latent variables. In order to predict in time we have to modify the way in which the GAN is used. When predicting with a GAN trained to produce $m$ time levels simultaneously, we need to know the first $m-1$ time levels in order to predict a future value, i.e.~from known solutions at time levels $\{0, 1, \cdots, m-2\}$ we can predict the solution at time level $m-1$. To predict the next time level, we use known solutions at time levels from $\{1, 2, \cdots, m-2\}$ and the newly predicted solution at time level $m-1$, to predict the solution at time level $m$. 

Assume we have the solutions at time levels up to and including~$m-2$ for the POD coefficients, denoted by $\{\tilde{\bm{\alpha}}^{k}\}_{k=0}^{m-2}$, and consider model parameters known over the entire simulation time $\bm{\tilde\mu}^k$, then to predict future solutions: 
\begin{enumerate}
\item a latent vector~$^{(0)} \mathbf{z}^{m-1}$ is randomly generated in order to start the prediction of time level $m-1$. The superscript in brackets on the left of the latent vector is the optimisation iteration counter within a time step prediction;
\item time iteration counter is set to $n=m-1$;
\item optimisation iteration counter is set to $l=0$; 
\item the generator of the GAN is evaluated at the current value of the latent variables, $^{(l)} \mathbf{z}^{n}$, yielding
\begin{equation}
G(^{(l)} \mathbf{z}^{n}) = \: ^{(l)}\bm{\Phi}^{n};
\end{equation}
\item the difference between the predicted values and the known values is calculated:
\begin{multline}\label{eq:loss_prediction}
\mathcal{L}_p(^{(l)}\mathbf{z}^{n}) =  \sum_{k=n-m+1}^{n-1}\, \left(\tilde{\bm{\alpha}}^k - ^{(l)}\!\!\bm{\alpha}^k \right)^T \bm{W}_\alpha \left(\tilde{\bm{\alpha}}^k - ^{(l)}\!\!\bm{\alpha}^k \right)  \\
+  \sum_{k=n-m+1}^{n-1}\,  \zeta_\mu\left(\tilde{\bm{\mu}}^k - ^{(l)}\!\!\bm{\mu}^k \right)^T \bm{W}_\mu \left(\tilde{\bm{\mu}}^k - ^{(l)}\!\!\bm{\mu}^k \right), 
\end{multline} 
where $\bm{W}_\alpha$ is a square matrix of size $N_{\text{POD}}$ whose diagonal values are equal to the weights that govern the relative importance of the POD coefficients. All other entries are zero. The weights could be based on the singular values if a POD method is used for compression, for example. $\bm{W}_\mu$ is a square matrix of size $N_{\mu}$ whose diagonal values are equal to the model parameter weights, and the scalar $\zeta_\mu$ controls how much importance is given to the model parameters compared to the compressed variables. It is worth mentioning that the goal in each time iteration is to predict a new time level $n$, hence the POD coefficients ${\bm{\alpha}}^n$ and model parameters ${\bm{\mu}}^n$ of this time level are not in the loss function.
\item the gradient of the loss $\mathcal{L}_p$ is  calculated with respect to the latent variables~$^{(l)} \mathbf{z}^n$ (by back-propagation), and $\mathcal{L}_p$ is minimised in the gradient direction leading to an updated set of latent variables~$^{(l+1)} \mathbf{z}^n$;
\item the optimisation iteration counter is incremented by one ($l \leftarrow l+1$); 
\item steps~4 to~7 are repeated until convergence is reached;
\item the converged latent variables are saved as $\mathbf{z}^n$ (note, no optimisation iteration index) and used to initialise the latent variables at the next time level, $^{(0)}\mathbf{z}^{n+1} = \mathbf{z}^{n}$. The predicted time step $n$ is added to the known solutions $\tilde{\bm{\alpha}}^{n}$ = $\bm{\alpha}^{n}$;
\item the time iteration counter is incremented by one ($n \leftarrow n+1$);
\item go back to step 3 (until the final time level is reached).
\end{enumerate}

\begin{figure}[htb]
	\centering
	\includegraphics[width=0.85\linewidth]{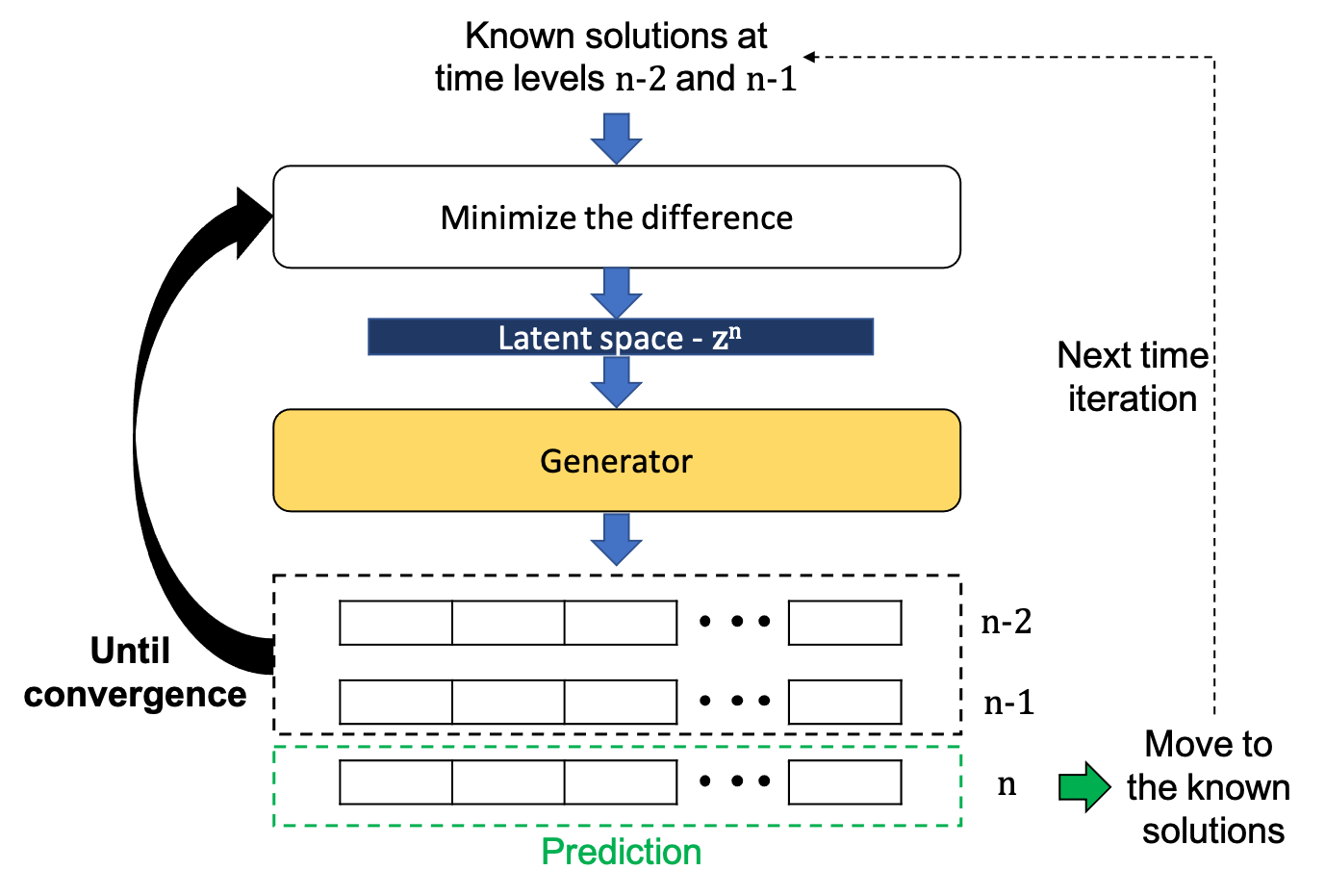}
	\caption{One time iteration of the PredGAN for a sequence of three time levels ($m=3$).}
	\label{fig:PRGANc}
\end{figure}

It is worth mentioning that the gradient of Eq.~\eqref{eq:loss_prediction} can be calculated by automatic differentiation \citep{wengert:64,linnainmaa:76,baydin:17}. In other words, minimising the loss in Eq.~\eqref{eq:loss_prediction} can be achieved simply by back-propagating the loss through the generator using the same methods that were employed when training the GAN. Figure~\ref{fig:PRGANc} illustrates how the PredGAN works for a generator trained to produce a sequence of 3 time steps (i.e. $m=3$). One important aspect of this predictive GAN approach to time stepping is that it never tries to extrapolate, only interpolate previous data. Thus the results of this model will always look realistic if the GAN is well trained and every point in the latent space $\mathbf{z}$ produces realistic looking models.   

\subsection{Data assimilation using GANs}
\label{sec:data_assimilation}

Data assimilation is an inverse problem that aims to combine a mathematical model with observations \citep{tarantola:05,oliver:08}. Data assimilation can be executed naturally by GANs due to their inherent adjoint-like nature. To perform data assimilation with GANs, we propose a method named Data Assimilation Predictive GAN (DA-PredGAN) that performs the following changes in the prediction algorithm (PredGAN). First, one additional term is included in the functional (loss) in Eq.~\eqref{eq:loss_prediction} to account for the data mismatch between the generated values and observations. Secondly, instead of knowing the model parameters $\bm{\mu}^k$, as in the prediction, for the data assimilation the goal is to match the observed data and determine the values of $\bm{\mu}^k$. Thirdly, the forward marching in time is now replaced by forward and backward marching.
 
\subsubsection{The functional for a time level} 

Analogous to the definition of $\Phi$ in equation~\eqref{phi}, the primitive variables vector and observations vector are defined as
\begin{equation}
\bm{u}^k  = \left( \begin{array}{c} u_1^k \\ u_2^k \\ \vdots \\ u_{N_c}^k \\ \end{array} \right) \ , \qquad 
\left(\bm{u}^{\text{obs}}\right)^k  = \left( \begin{array}{c} (u^{\text{obs}})_1^k \\ (u^{\text{obs}})_2^k \\ \vdots \\ (u^{\text{obs}})_{N_c}^k \\ \end{array} \right)\,
\end{equation} 
in which the primitive variable at node $j$ of the spatial grid and time level $k$ is $u_j^k$, and the observation at node $j$ of the spatial grid and time level $k$ is $(u^{\text{obs}})_j^k$. $N_c$ is the number of nodes or cells in the grid. For the forward march, we perform the same process as the prediction in time (Section \ref{sec:prediction_in_time}); however, the functional for time level $n$ is now written as
\begin{multline}\label{eq:loss_da_forward}
\mathcal{L}_{da,f}(\mathbf{z}^{n}) =  \sum_{k=n-m+1}^{n-1}\, \left(\tilde{\bm{\alpha}}^k - \bm{\alpha}^k \right)^T \bm{W}_\alpha \left(\tilde{\bm{\alpha}}^k - \bm{\alpha}^k \right)  \\
+  \sum_{k=n-m+1}^{n-1}\,  \zeta_\mu\left(\tilde{\bm{\mu}}^k - \bm{\mu}^k \right)^T \bm{W}_\mu \left(\tilde{\bm{\mu}}^k - \bm{\mu}^k \right) \\
+  \sum_{k=n-m+1}^{n-1}\,  \zeta_{obs}\left(\bm{u}^k - (\bm{u}^{\text{obs}})^k \right)^T \bm{W}_u^k \left(\bm{u}^k - (\bm{u}^{\text{obs}})^k \right), 
\end{multline} 
where $\bm{W}_u^k$ is a square matrix of size $N_c$ whose diagonal values are equal to the observed data weights, and the scalar $\zeta_{obs}$ direct controls how much importance is given to the data mismatch. The values in the diagonal of $\bm{W}_u^k$ are set to zero where we have no observation. The subscript $f$ of $\mathcal{L}_{da}$ indicates that this loss function applies to the forwards march.

In the previous section, before the GAN can start predicting in time, it requires $m-1$ known solutions of the POD coefficients $\{\tilde{\bm{\alpha}}^{k}\}_{k=0}^{m-2}$ corresponding to the first $m-1$ time levels, and the model parameters $\bm{\tilde\mu}^k$ over all simulation time. There are two sets of POD and model variables: a set of known variables $(\bm{\tilde{\alpha}}^k, \bm{\tilde\mu}^k)$ and a set of predicted values $(\bm{\alpha}^k, \bm{\mu}^k)$. When assimilating data, we usually do not know the values of the model parameters, hence the aim is to match the observed data and determine the corresponding $\bm{\mu}^k$. To this end, during a time iteration of the forward and backward marches the known variables of the model parameters are updated by the newly predicted time step $\bm{\tilde\mu}^n = \bm{\mu}^n$, the same way as for the POD coefficients (item 9 of the prediction process, Section \ref{sec:prediction_in_time}). This gives the best approximation to these variables and allows them to vary during the data assimilation process. Furthermore, after the forward march the solutions at the last $m-1$ time levels are used as known solutions to start the backward march, and after a backward march the solutions at the first $m-1$ time levels are used as known solutions to start the next forward march. 

For marching backwards in time, the loss function should be modified thus
\begin{multline}\label{eq:loss_da_backward}
\mathcal{L}_{da,b}(\mathbf{z}^{n}) = \sum_{k=n+1}^{n+m-1}\, \left(\tilde{\bm{\alpha}}^k - \bm{\alpha}^k \right)^T \bm{W}_\alpha \left(\tilde{\bm{\alpha}}^k - \bm{\alpha}^k \right)  \\
+  \sum_{k=n+1}^{n+m-1}\,  \zeta_\mu\left(\tilde{\bm{\mu}}^k - \bm{\mu}^k \right)^T \bm{W}_\mu \left(\tilde{\bm{\mu}}^k - \bm{\mu}^k \right) \\
+  \sum_{k=n+1}^{n+m-1}\,  \zeta_{obs}\left(\bm{u}^k - (\bm{u}^{\text{obs}})^k \right)^T \bm{W}_u^k \left(\bm{u}^k - (\bm{u}^{\text{obs}})^k \right), 
\end{multline}
where the subscript $b$ of $\mathcal{L}_{da}$ indicates that this loss function applies to the backwards march.
It is worth noting that the only difference between  Eq.~\eqref{eq:loss_da_forward} and Eq.~\eqref{eq:loss_da_backward} is the indices in the summation. For the backward march, given the solution at time levels $\{n+m-1, n+m-2, \cdots, n+1\}$, we can predict the solution at time level $n$. Then, to predict the next time level we use known solutions at time levels from $\{n+m-2, n+m-3, \cdots, n+1\}$ and the newly predicted solution at time level $n$, to predict the solution at time level $n-1$. We continue the process until predicting the first time step. After the backward march, we calculate the average data mismatch, between the predicted primitive variables and the observed data (last term on the right of Eqs.~\eqref{eq:loss_da_forward} and~\eqref{eq:loss_da_backward}), through all the last backward and forward iterations. The process continues with a new forward and backward march until the average mismatch has converged or the maximum number of forward-backward iterations is reached.

\subsubsection{Observations of the primitive variables} 

If proper orthogonal decomposition is used to compress the grid variables, in which 
\begin{equation}
\bm{u}^k =  \bm{B}\bm{\alpha}^k + \overline{\bm{u}},
\end{equation} 
a functional contribution that can be used directly within the optimiser is 
\begin{equation}
\mathcal{L}_{obs}(\mathbf{z}^{n}) = \sum_{k}\,  \zeta_{obs}\left(\bm{B}\bm{\alpha}^k + \overline{\bm{u}} - (\bm{u}^{\text{obs}})^k \right)^T \bm{W}_u^k \left(\bm{B}\bm{\alpha}^k + \overline{\bm{u}} - (\bm{u}^{\text{obs}})^k \right), 
\end{equation} 
where the columns of the matrix $\bm{B}$ are the basis functions which relate the high-dimensional solution variables to the POD coefficients, and $\overline{\bm{u}}$ is the mean of the ensemble of snapshots for the variable $\bm{u}$.

Having written the solution in terms of the compressed variables when calculating the mismatch of the observations, the final version of the functional for the forward and backward march is

\begin{multline}
\mathcal{L}_{da}(\mathbf{z}^{n}) = \sum_{k}\, \left(\tilde{\bm{\alpha}}^k - \bm{\alpha}^k \right)^T \bm{W}_\alpha \left(\tilde{\bm{\alpha}}^k - \bm{\alpha}^k \right) 
+  \sum_{k}\,  \zeta_\mu\left(\tilde{\bm{\mu}}^k - \bm{\mu}^k \right)^T \bm{W}_\mu \left(\tilde{\bm{\mu}}^k - \bm{\mu}^k \right) \\
+  \sum_{k}\,  \zeta_{obs}\left(\bm{B}\bm{\alpha}^k + \overline{\bm{u}} - (\bm{u}^{\text{obs}})^k \right)^T \bm{W}_u^k \left(\bm{B}\bm{\alpha}^k + \overline{\bm{u}} - (\bm{u}^{\text{obs}})^k \right),
\end{multline} 

where for the forward march $k \in \{ n-m+1, n-m+2, \cdots, n-1\}$ and for the backward march $k \in \{ n+m-1, n+m-2, \cdots, n+1\}$.

\subsubsection{Applying relaxation} 

To stabilise the process of marching forwards and backwards, a relaxation parameter is used in the resulting latent vector $\mathbf{z}^n$ at each time iteration. After performing an optimisation using Eqs.~\eqref{eq:loss_da_forward} or \eqref{eq:loss_da_backward}, the resulting latent vector is relaxed by 
\begin{equation}\label{eq:relax}
    \mathbf{z}^n = (1-r^j)\mathbf{z}^{n-1} + r^j\mathbf{\hat{z}}^{n}
\end{equation}
where $\mathbf{\hat{z}}^{n}$ is the resulting latent vector generated in each time iteration by minimising Eqs.~\eqref{eq:loss_da_forward} or \eqref{eq:loss_da_backward}. $j$ represents a iteration corresponding to a entire forward and backward march, and $r^j$ is the relaxation factor used in the iteration $j$. Thus each relaxation parameter $r^j$ is used in all time iterations $n$ within the forwards and backwards march.   

The relaxation factor $r^j$ starts the data assimilation process with the value one. If the average data mismatch at $j$ is greater than at $j-1$, then $r^j = r^j/2$ and the forward and backward iteration $j$ is repeated. On the other hand, if the average data mismatch at $j$ is less than at $j-1$, the algorithm goes to the next iteration $j+1$ using  $r^{j+1} = 1.5 r^j$, also respecting the maximum value of one for the relaxation factor (if $r^{j+1} > 1$ then $r^{j+1} = 1$).  

\subsubsection{Data assimilation algorithm through time} 
\label{sec:dattime}

To assimilate data through time the following steps are performed:
\begin{enumerate}
\item march forward in time using the Eq.~\eqref{eq:loss_prediction} and the prediction process in Section \ref{sec:prediction_in_time}, with guessed parameters $\{\tilde{\bm{\alpha}}^{k}\}_{k=0}^{m-2}$ for the first $m-1$ time steps and $\bm{\tilde\mu}^{k}$ over all simulation time. This will results in an initial guess of $\bm{\alpha}^n$ at all time levels $n$.

\item Time march backwards in time optimising  Eq.~\eqref{eq:loss_da_backward}, starting from the $m-1$ final time levels (obtained from the previous iteration). This tries to perform time stepping while attempting to match the observations using the observed data mismatch part of the functional (last term on the right of Eq.~\eqref{eq:loss_da_backward}). During the time march update the model parameters using the newly predicted time step $\bm{\tilde\mu}^n = \bm{\mu}^n$.

\item Keep time stepping forwards till the end of time and then backwards to the start of time using Eqs.~\eqref{eq:loss_da_forward}, \eqref{eq:loss_da_backward} and \eqref{eq:relax}, until the algorithm has converged. We use as convergence criteria $r^j < 0.01$.    

\item If there are parameters $\bm{\alpha}^n$ and $\bm{\mu}^n$ changing rapidly during the data assimilation process increase the corresponding weight (in Eqs.~\eqref{eq:loss_da_forward} and \eqref{eq:loss_da_backward}) and if they are not changing rapidly enough decrease it. 

\item After convergence, perform a last forward march using the Eq.~\eqref{eq:loss_prediction} and the prediction process in Section \ref{sec:prediction_in_time}. Use the last calculated parameters $\{\tilde{\bm{\alpha}}^{k}\}_{k=0}^{m-2}$ and $\bm{\tilde\mu}^k$ for this.  
\end{enumerate}

\begin{figure}[htb]
	\centering
	\includegraphics[width=0.6\linewidth]{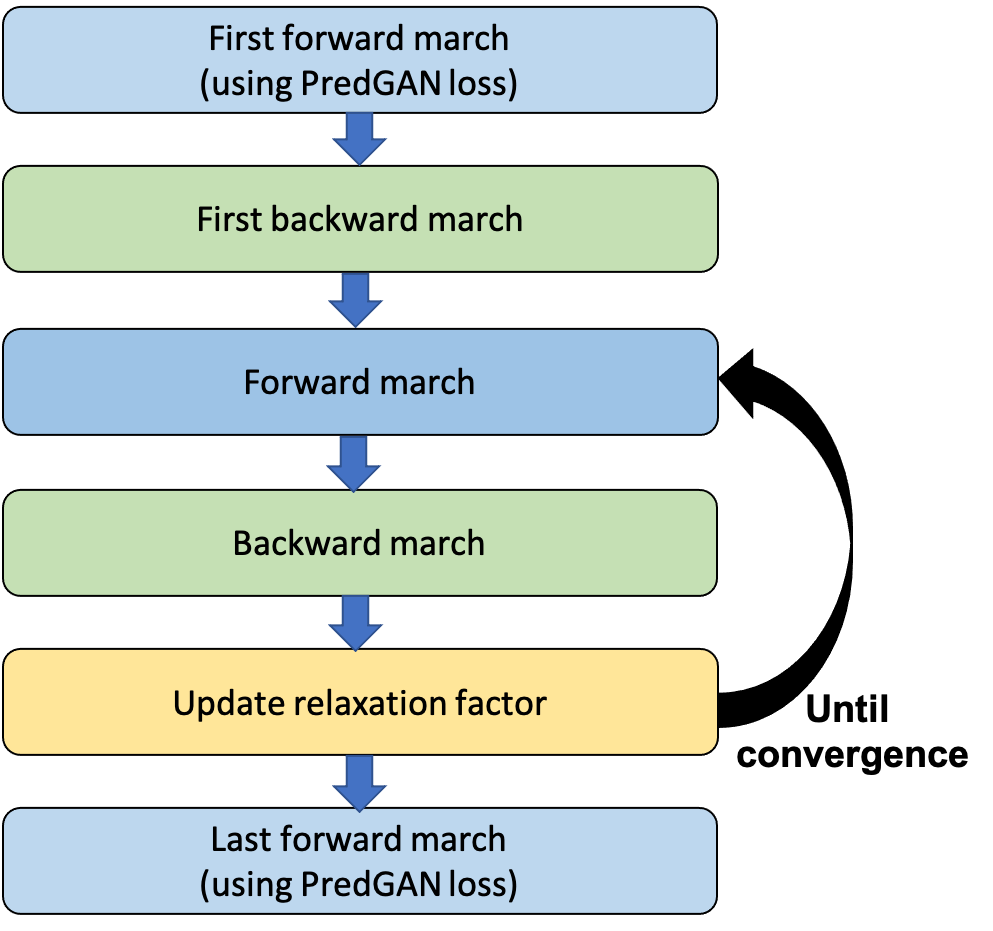}
	\caption{Overview of the DA-PredGAN process.}
	\label{fig:DA-PredGAN}
\end{figure}

Figure \ref{fig:DA-PredGAN} shows an overview of the DA-PredGAN algorithm.

\subsection{Weighting terms in the functionals}\label{sec:weighting_terms}

We suggest giving the data mismatch part of the functional greater priority than the time stepping part of the functional. Considering that we set non-zero terms on the diagonal of $\bm{W}_u^k$ to one, then 
\begin{equation}
\zeta_{obs}=  \hat\zeta_{obs} \left(  \frac{ \Delta\alpha } {\Delta u } \right)^2
\left(  \frac{  (m-1) \sum_{i=1}^{N_{\text{POD}}} (w_\alpha)_{ii}}{\sum_{k} \sum_{i=1}^{N_c}  (w_u)_{ii}^k }  \right) , 
\end{equation}
where $\hat\zeta_{obs}$ is a tuning parameter and in this work it is set to 10. $\Delta\alpha$ and $\Delta u$ are the ranges of the compressed variables and the primary variables, respectively. $(w_\alpha)_{ii}$ are the the terms on the diagonal of $\bm{W}_\alpha$, and  $(w_u)_{ii}^k$ are the terms on the diagonal of $\bm{W}_u^k$. For the forward march $k \in \{ n-m+1, n-m+2, \cdots, n-1\}$ and for the backward march $k \in \{ n+m-1, n+m-2, \cdots, n+1\}$.

$\zeta_\mu$ controls how quickly one lets the parameters $\bm{\mu}$ change within the data assimilation method. We choose to set all the terms on the diagonal of $\bm{W}_\mu$ to one, and 
\begin{equation} 
\zeta_\mu = \hat\zeta_\mu \left( \frac{\Delta\alpha}{\Delta\mu}\right)^2  \left(  \frac{ \sum_{i=1}^{N_{\text{POD}}} (w_\alpha)_{ii}}{\sum_{i=1}^{N_\mu}  (w_\mu)_{ii} }  \right) , 
\label{gamma-def} 
\end{equation}
where ${\Delta\mu}$ represents the range of the scalar parameters, $(w_\mu)_{ii}$ are the the terms on the diagonal of $\bm{W}_\mu$, and $\hat\zeta_\mu$ is a tuning parameter. In this work, we use $\hat\zeta_\mu = 10^{-2}$ for the prediction, and the first and last forward marches of the data assimilation. For the other forward and backward marches of the data assimilation we can choose to dynamically update $\zeta_\mu$, in order to let $\bm{\mu}^k$ change more rapidly at the beginning and more slowly when the process is near convergence. Therefore, we start with $\hat\zeta_\mu = 10^{-4}$ and increase it by a factor of $1.2$ after each forward-backward iteration.         

\section{Test case}
\label{sec:testcase}

\subsection{Compartmental models in epidemiology}
\label{sec:epid}

The current COVID-19 pandemic, caused by the virus SARS-CoV-2, is something without precedent in modern history, although it follows the same rules common to other pathogens \citep{cobey:20}. The knowledge gathered during more than one century studying these outbreaks has given rise to a well-founded theory of the dynamics of infectious diseases \citep{anderson:92,bjornstad:18}. One of the simplest nonlinear models to describe the spread of an infection is the SIR model, which consists of a system of ordinary differential equations where S, I and R represent the number of people who are susceptible, infectious or recovered \citep{anderson:92, bjornstad:20} referred to as compartments. 

The model starts by considering a population of $N$ individuals in the susceptible compartment. If one individual with the disease is introduced into the population, over time, other people will become infected and move into the infectious compartment. The members of the infectious compartment will spread the pathogen among the population until they recover. This is called a ``closed epidemic'' for which $N = S + I + R$. The SIR model assumes that the population mixes at random and that it is large enough for averages to be used meaningfully \citep{bjornstad:18,bjornstad:20}. One important factor in the dynamics of infectious diseases, and consequently in the SIR model, is the basic reproduction number ($\mathcal{R}_0$) \citep{hethcote:00}. It represents the expected number of secondary cases caused by a single infected member in a completely susceptible population. Knowing the magnitude of the $\mathcal{R}_0$ gives an indication of how rapidly the infection could spread, allowing governments and authorities to estimate the amount of effort necessary to prevent, diminish or eliminate an infection from a population \citep{dietz:93}. Following this line, much effort has been committed to estimating the $\mathcal{R}_0$ in the COVID-19 pandemic worldwide \citep{liu:20,flaxman:20,shim:20,breda:20}. % JSC       

Albeit simple, the SIR model can provide important  insights into the dynamics of infectious diseases in an idealised population. Nonetheless, for more realistic situations other factors need to be taken into account such as births, deaths and loss of immunity. Furthermore, it is well known for most diseases that there is an incubation period between being infected and becoming infectious \citep{bjornstad:20b}. For that reason the SIR model can be extended to the SEIRS model. In the latter formulation, after being infected an individual is moved to the exposed compartment (E) and remains there until they become infectious. Also, recovered people may become susceptible again due to the loss of immunity. Another important factor regarding the flow in and out of a compartment is demography. Births and deaths can also be taken into account by adding their rates to the formulations \citep{bjornstad:20b}. Other types of compartments can also be added depending on the flow patterns between the compartments \citep{hethcote:00,Radulescu:20}. 

When large-scale simulations are considered, for example a simulation of spatial-variation of the COVID-19 infection in a whole city or a country, the computational time becomes a concern. Additionally, if observed data needs to be taken into account the whole process can become impracticable. In order to tackle this problem, we propose a surrogate model that can be used to replace the forward numerical simulation and it can also assimilate data without any additional run of the high-fidelity model. 

%\citep{safta:11, vyasarayani:20}

\subsubsection{SEIRS model}
\label{subsec:seirs}

\begin{figure*}[!htb]
	\centering
	\includegraphics[width=0.9\textwidth]{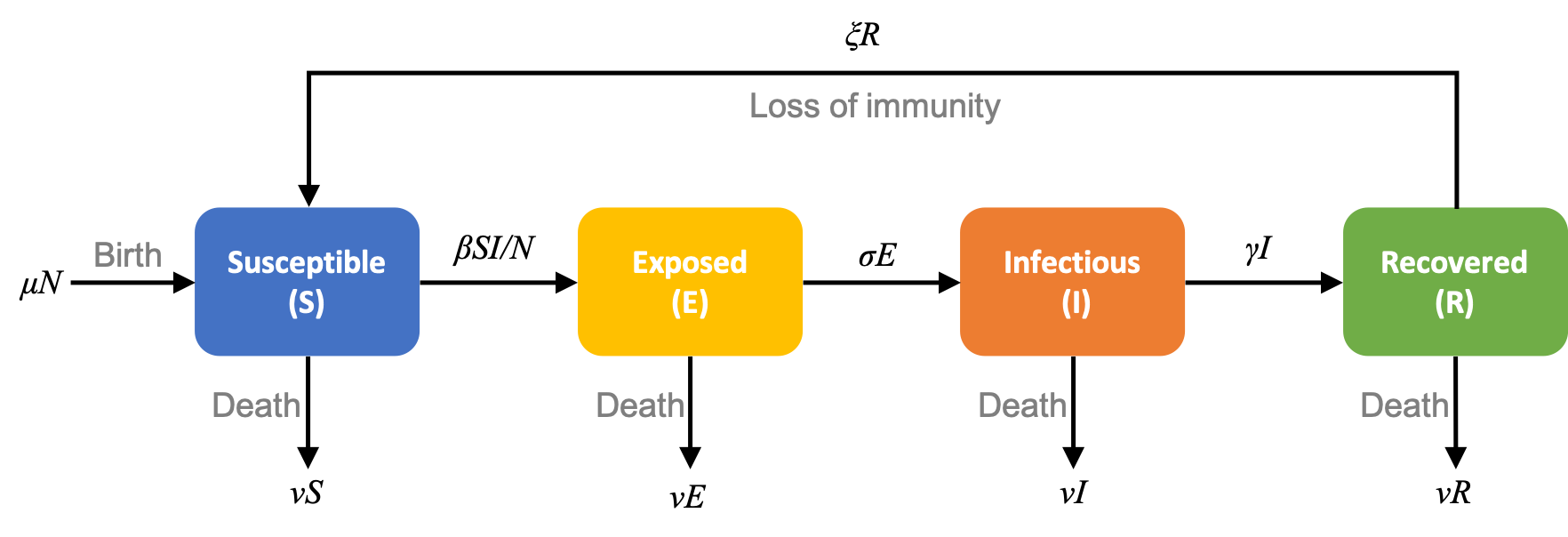}
	\caption{Diagram of the SEIRS model. It represents the number of people in each compartment (susceptible, exposed, infectious, recovered) and how they move between them.}
	\label{fig:SEIRSmodel}
\end{figure*}

The classic SEIRS (Susceptible - Exposed - Infectious - Recovered - Susceptible) model can be represented by the diagram in Figure \ref{fig:SEIRSmodel}. The diagram shows how individuals move from one compartment to another.   

In the diagram, $S$, $E$, $I$ and $R$ are the number of individuals in the susceptible, exposed, infectious and recovered compartments, respectively. $N$ represents the total population size, $\beta$ is the transmission rate (the average rate at which an infectious individual can infect a susceptible), $\sigma$ is the rate of exposed individuals becoming infectious ($1/\sigma$ is the average period in the exposed group), $\gamma$ is the recovered rate ($1/\gamma$ is the average infectious period), and $\xi$ is the rate recovered individuals return to the susceptible group ($1/\xi$ is the average period before loss of immunity). The vital dynamics are represented by $\eta$ and $\nu$, where $\eta$ is the birth rate and $\nu$ is the death rate.

The system of differential equations describing the SEIRS model dynamics can be expressed as 
\begin{subequations} \label{SEIRS-eqn}
	\begin{eqnarray}
	\frac{d S}{d t} & = &\eta N  - \frac{\beta SI}{N} + \xi R - \nu S, \\
	\frac{d E}{d t} & = &\frac{\beta SI}{N} - \sigma E - \nu E, \\ 
	\frac{d I}{d t} & = &\sigma E - \gamma I - \nu I, \\
	\frac{d R}{d t} & = &\gamma I  - \xi R- \nu R,
	\end{eqnarray}
\end{subequations}
where each equation represents the dynamics within a compartment \citep{bjornstad:20b}. At time $t$, the total number of people can be expressed as $N(t) = S(t) + E(t) + I(t) + R(t)$. If the birth rate is equal to the death rate ($\eta=\nu$) the total size of the population ($N$) remains constant over time. For this case the associated basic reproduction number is defined as 
\begin{equation}
    \label{eq:R0}
    \mathcal{R}_0 = \frac{\sigma}{(\sigma + \nu)}\frac{\beta}{(\gamma+\nu)}.
\end{equation}
For most acute infections, the death rate $\nu$ is much smaller than the epidemic rates, thus in realistic situations it barely affects the evolution of the disease \citep{bjornstad:20b}. 

Eqs.~\eqref{SEIRS-eqn} can be also expanded into the extended SEIRS compartmental equations to take into account the spatial variation of the disease \citep{quilodran:21}. 

\subsubsection{Extended SEIRS model}
\label{sec:eseirs}

People movement is of paramount importance to model the spread of infection diseases such as COVID-19. Therefore, this project extends Eqs.~\eqref{SEIRS-eqn} in two ways: 
\begin{enumerate}
    \item It includes two people groups, people at home ($h=1$) and people who are mobile ($h=2$);
    \item It applies transport via diffusion to model the movement of people through the domain.
\end{enumerate}

As a result, the extended equations could model the daily cycle of night and day for the transient calculations, in which there is a ``pressure'' for mobile people to return to their homes at night and join the home group, and a similar pressure for people to leave their homes during the day, who will therefore join the mobile group. To accomplish this, the extended SEIRS model introduces a diffusion term (last term on the right of Eqs.~\eqref{eq:Ext-SEIRS}) and an interaction term (penultimate term on the right of Eqs.~\eqref{eq:Ext-SEIRS}) to model this process: 

\begin{subequations}
\label{eq:Ext-SEIRS} 
\begin{align} 
%\begin{eqnarray}
\frac{\partial S_h}{\partial t} & = \eta_h N_h  - \frac{S_h \sum_{h'}(\beta_{h\, h'}   I_{h'}) }{ N_{h}} 
+  \xi_h R_h - \nu_h^S S_h - \sum_{h'=1}^{\cal H} \lambda_{h\,h'}^S S_{h'} 
+ \nabla \cdot ( k_h^S \nabla S_h  ) , 
\label{eq:Ext-SEIRS-eq1} \\[3mm]%\end{eqnarray} 
%\begin{eqnarray}
 \frac{\partial E_h}{\partial t}  & =  \frac{S_h \sum_{h'}(\beta_{h\, h'}   I_{h'}) }{ N_{h}}
-  \sigma_h E_h - \nu_h^E E_h  
- \sum_{h'=1}^{\cal H} \lambda_{h\,h'}^E E_{h'}  + \nabla \cdot ( k_h^E \nabla E_h ),
\label{eq:Ext-SEIRS-eq2}
\\[2mm]%\end{eqnarray}  
%\begin{eqnarray}
\frac{\partial I_h}{\partial t}  & =   \sigma_h E_h - \gamma_h I_h - \nu_h^I I_h 
- \sum_{h'=1}^{\cal H} \lambda_{h\,h'}^I I_{h'} 
+ \nabla \cdot (k_h^I \nabla I_h ),
\label{eq:Ext-SEIRS-eq3} 
\\[2mm]%\end{eqnarray}
%\begin{eqnarray}
\frac{\partial R_h}{\partial t}  & =   \gamma_h I_h  - \xi_h R_h 
- \nu_h^R R_h
- \sum_{h'=1}^{\cal H} \lambda_{h\,h'}^R R_{h'}  
+ \nabla \cdot ( k_h^R \nabla R_h ), 
\label{eq:Ext-SEIRS-eq4} 
%\end{eqnarray}
\end{align}
\end{subequations} 
in which  ${\cal H}$ represents the number of people and/or places groups e.g.~people at home, mobile people, people at work in the office, shops,  people in hospital. Here, we have two groups of people, hence ${\cal H}=2$, one representing people at home, $h=1$, and the second representing people that are mobile and outside their homes therefore, $h=2$. The diffusion coefficient is represented by $k_h$ and  describes the movement of people around the domain. It is defined for each compartment denoted by a superscript \{$S, E, I, R$\}. In addition, $\beta_{h\, h'}$ determines not only the transmission rate between compartments, but also how the disease is transmitted from people in group $h'$ to people in group $h$. The interaction terms, $\lambda_{h\, h'}$, control how people move between groups, for example, how people that are in the mobile group goes to the home group. When moving between groups people remain in the same compartment, and when moving between compartments, people remain in the same group. 

If we again consider the same value for the birth and the death rate in all groups, the total size of the population ($N = \sum_h N_h$) remains constant over time. For this case the associated basic reproduction number for each group is defined as 
\begin{equation}
    \label{eq:R0ext}
    \mathcal{R}_{0\,h} = \frac{\sigma_h}{(\sigma_h + \nu_h)}\frac{\beta_{h\,h}}{(\gamma_h+\nu_h)},
\end{equation}
where we assume $\beta_{h\,h'}=0$ when $h \not= h'$ because the people occupying these groups never meet i.e.~people in their homes never meet mobile people (who are outside their homes). 

\begin{figure*}[!htb]
	\centering
	\includegraphics[width=0.9\textwidth]{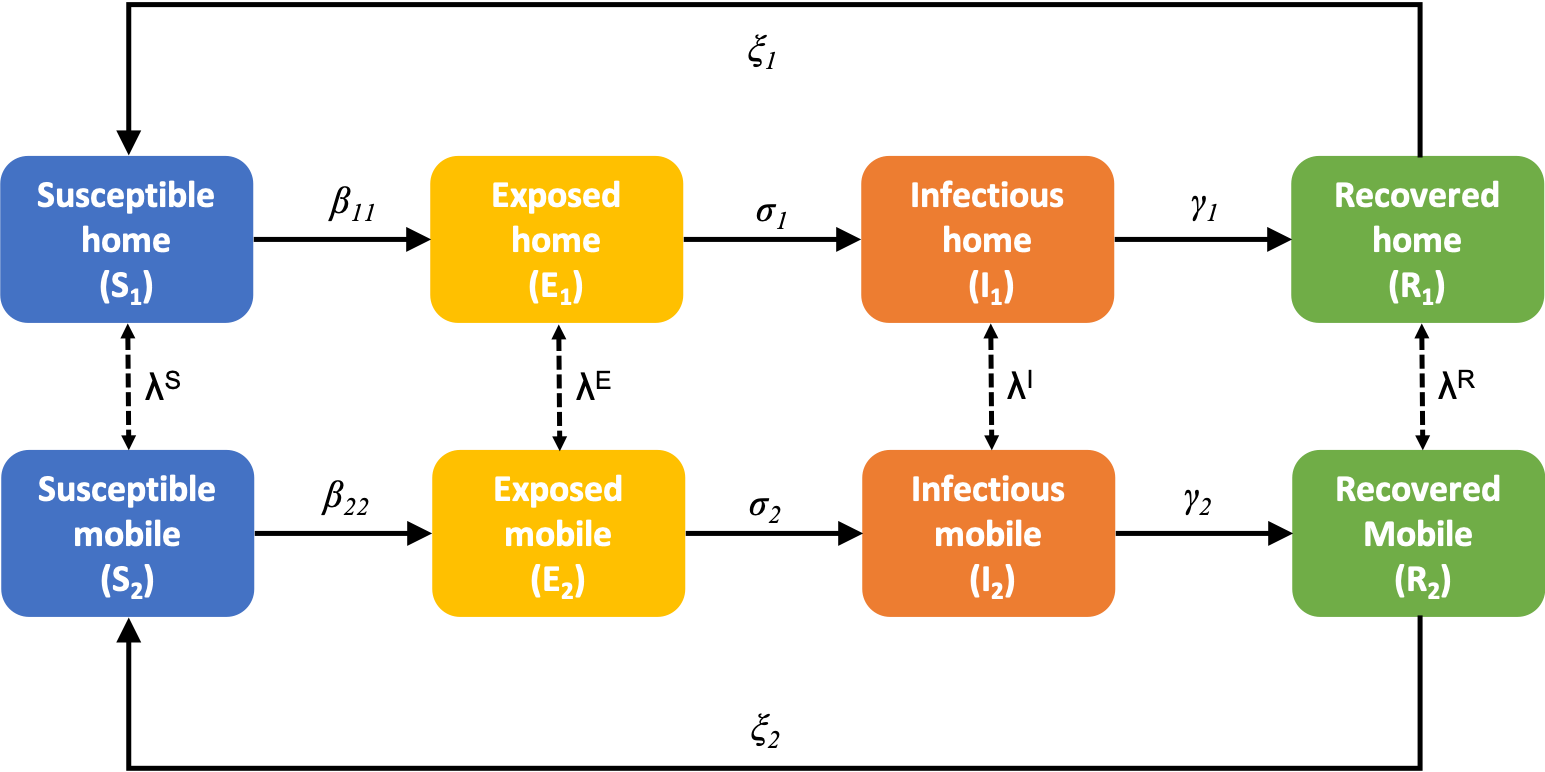}
	\caption{Diagram of the extended SEIRS model for one point in space (or one cell in the grid). The diagram shows how people move between groups and compartments within the same cell in the grid. The vital dynamics and the transport via diffusion is not displayed here.}
	\label{fig:ExtSEIRSmodel}
\end{figure*}

It is worth mentioning that instead of having the number of people in each compartment ($S$, $E$, $I$ and $R$) changing only in time, as in Eqs.~\eqref{SEIRS-eqn}, in the extended SEIRS model, Eqs.~\eqref{eq:Ext-SEIRS}, they can vary in space and time. Figure \ref{fig:ExtSEIRSmodel} shows, for one position in space (or one cell in the grid), how people move between groups (home, and mobile) and compartments. Futher details about the extended SEIRS model can be found in \citet{quilodran:21}.  

\subsubsection{Discretisation and solution methods used} 

The spatial variation is discretised on a regular grid of $N_X \times N_Y \times N_Z$ control volume cells. Here we work on a 2D problem, so $N_Z=1$. We use a five-point stencil and second order differencing of the diffusion operator, as well as backward Euler time stepping. We iterate within a time step, using Picard iteration, until convergence of all non-linear terms and evaluate these non-linear terms at the future time level. To solve the linear system of equations we simply use forward backward Gauss-Seidel (FBGS) within each group (each of the variables $S_1$, $E_1$, $I_1$, $R_1$, $S_2$, $E_2$, $I_2$, $R_2$) until convergence and Block FBGS between groups to obtain overall convergence of the system. This solver is sufficient to solve the test problems presented here. 

\subsection{Problem set up of an idealised town} 

The test case occupies an area of 100km by 100km as shown in Figure \ref{fig:grid_domain}. We divided this area in 25 regions, where those labelled as 1 are regions to which people do not travel, the region labelled as 2 is where homes are located, and regions from 2 to 10 are where people in the mobile group can travel. Thus people in the home group stay in the region 2. The aim is that most people move from home to mobile group in the morning, travel to locations in regions 2 to 10, and return to the home group later on in the day. The extended SEIRS model is used here to model this movement of people around the the cross-shaped domain in Figure \ref{fig:grid_domain}, in addition to calculating which compartment and group each person is in at a given time $t$. A person at any time and position within the domain belongs to one of the two groups, home or mobile, and is in one of the four compartments, Susceptible, Exposed, Infectious, or Recovered.      

\begin{figure}[htbp]
\centering
\begin{tikzpicture}

\filldraw[fill=gray!20!white, draw=green!40!black] (2,0) rectangle (3,5);
\filldraw[fill=gray!20!white, draw=green!40!black] (0,2) rectangle (2,3);
\filldraw[fill=gray!20!white, draw=gray] (3,2) rectangle (5,3);

\draw[step=1cm,color=gray, very thick] (0,0) grid (5,5);

\node at (0.5,+4.5) {1};
\node at (1.5,+4.5) {1};
\node at (2.5,+4.5) {6};
\node at (3.5,+4.5) {1};
\node at (4.5,+4.5) {1};

\node at (0.5,+3.5) {1};
\node at (1.5,+3.5) {1};
\node at (2.5,+3.5) {10};
\node at (3.5,+3.5) {1};
\node at (4.5,+3.5) {1};

\node at (0.5,+2.5) {3};
\node at (1.5,+2.5) {8};
\node at (2.5,+2.5) {4};
\node at (3.5,+2.5) {9};
\node at (4.5,+2.5) {5};

\node at (0.5,+1.5) {1};
\node at (1.5,+1.5) {1};
\node at (2.5,+1.5) {7};
\node at (3.5,+1.5) {1};
\node at (4.5,+1.5) {1};

\node at (0.5,+0.5) {1};
\node at (1.5,+0.5) {1};
\node at (2.5,+0.5) {2};
\node at (3.5,+0.5) {1};
\node at (4.5,+0.5) {1};

\end{tikzpicture}
\caption{Domain of the 100km $\times$ 100km idealised town showing the different regions. Regions 
where people can travel are shown in grey.  }
\label{fig:grid_domain}
\end{figure}

Table \ref{tab:epid} shows the epidemiological parameters used in this test case. These values are chosen as they are representative of the COVID-19 infection. $T_{day}$ is the number of seconds in one day, and the transmission rates $\beta_{h\,h}$ are calculated based on the vales of the $\mathcal{R}_{0\,h}$. To generate the ensemble of simulations for the training and test data, the basic reproduction number for each group of people is sampled from a uniform distribution with interval $(0, 20)$. The $\mathcal{R}_0$ is the main parameter to control the evolution of infectious diseases, hence we choose to vary it in the ensemble. We also choose to use a wider range of variation for the $\mathcal{R}_{0\,h}$, in accordance with \citet{kochanczyk:20}, than the most common values estimated for the COVID-19 pandemic. The diffusion coefficients ($k$) used to model the spatial movement of people and the interaction terms ($\lambda$) are the same as those in \citet{quilodran:21}.         
\begin{table}[!h] 
	\caption{Parameters for the extended SEIRS model. $T_{day}$ is the number of seconds in one day.} \centering
    \begin{tabular}{| l | c | c |}
		\hline
		\textbf{Parameter} & \textbf{Home group} &  \textbf{Mobile group} \\
		\hline\hline
		Birth and death rate ($\eta = \nu$) &  \multicolumn{2}{|c|}{$(60 \times 365 \times T_{day})^{-1}$} \\
		\hline
		Loss of immunity rate ($\xi$) & \multicolumn{2}{|c|}{$(365 \times T_{day})^{-1}$} \\	
		\hline
		Exposed to infectious rate ($\sigma$) & \multicolumn{2}{|c|}{$(4.5 \times T_{day})^{-1}$}   \\			
		\hline
	    Recovery rate ($\gamma$) &  \multicolumn{2}{|c|}{$(7 \times T_{day})^{-1}$} \\
		\hline
		Reproduction number & $\mathcal{R}_{0\,1} \sim  U(0,20)$ &  $\mathcal{R}_{0\,2} \sim U(0,20)$ \\
		\hline
	\end{tabular}
	\label{tab:epid}
\end{table}

The simulations are run for $45.5$ days with a time step of $\Delta t=4000$ seconds. We use uniformly space $10\times10$ control volumes to discretise the domain in Figure \ref{fig:grid_domain}. Therefore, each region in Figure \ref{fig:grid_domain} comprises four control volumes. We start the simulation with 2000 people in each control volume of region 2 and belonging to the home group. All other fields are set to zero. The initial condition is that $0.1\%$ of people at home have been exposed to the virus and will thus develop an infection.

\section{Results}
\label{sec:results}

\begin{figure*}[!htb]
	\centering
	\begin{subfigure}[t]{0.49\textwidth}
		\centering
		\includegraphics[width=\textwidth]{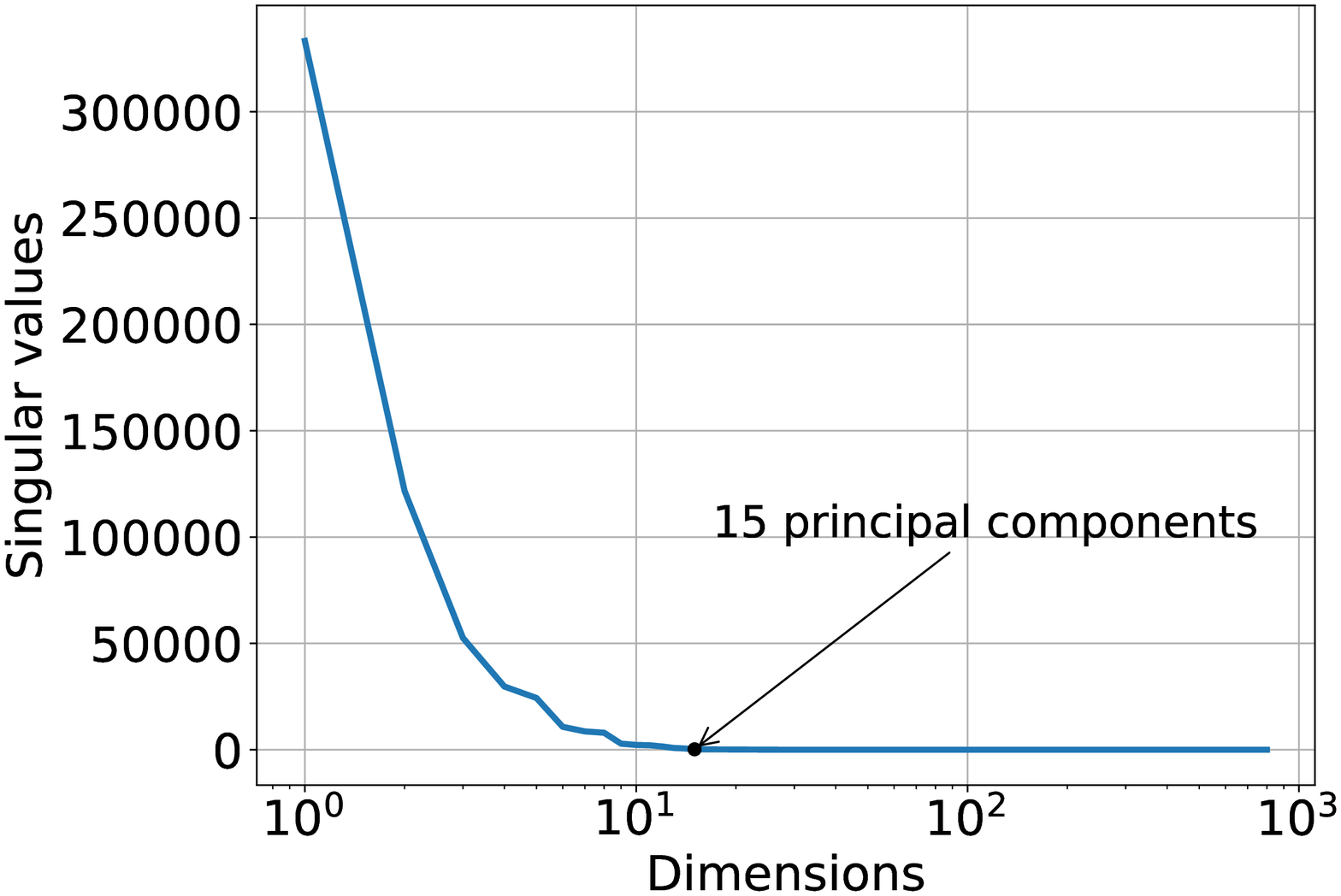}
		\caption{}
		\label{fig:pod_sv1}
	\end{subfigure}
	\begin{subfigure}[t]{0.49\textwidth}
		\centering
		\includegraphics[width=\textwidth]{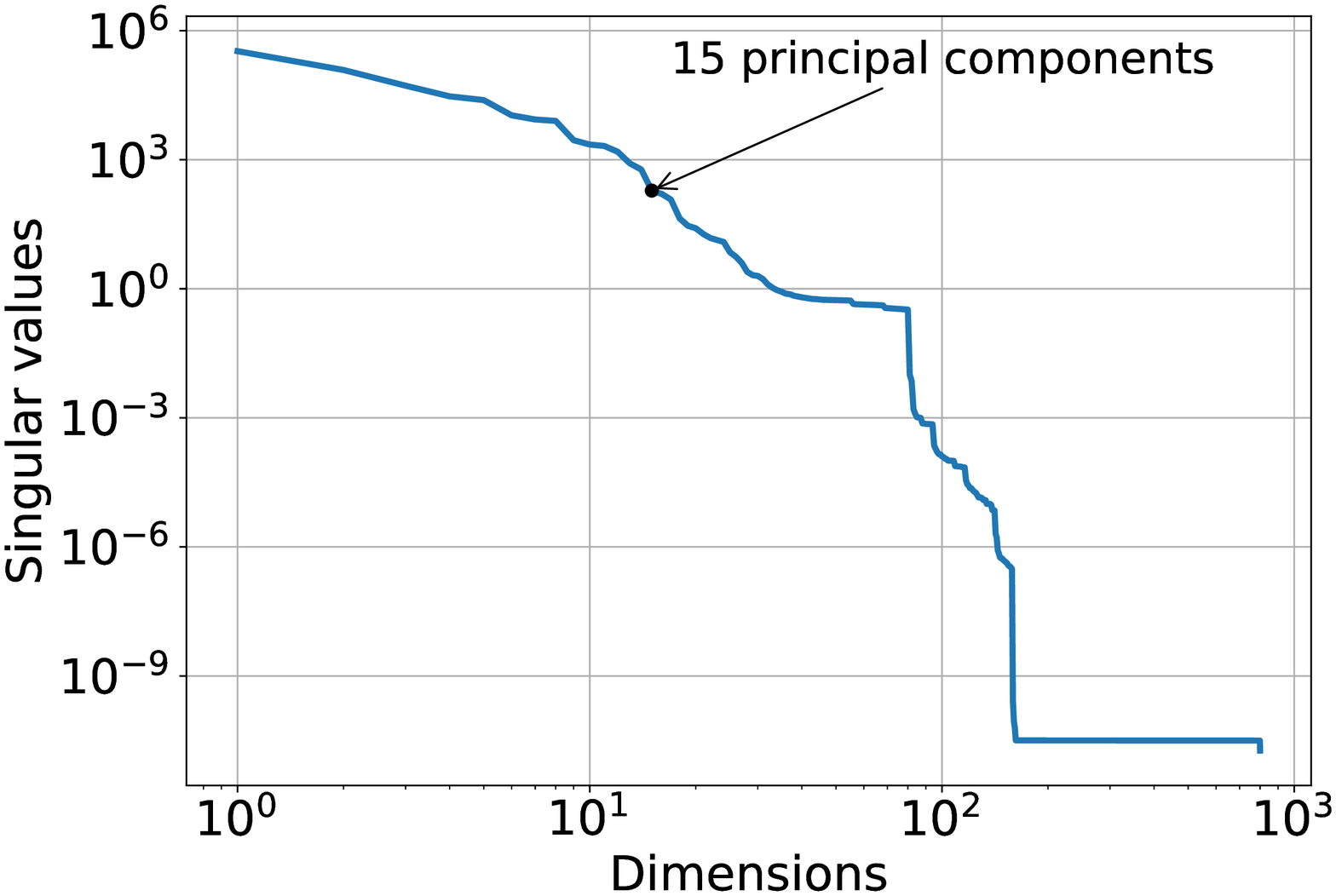}
		\caption{}
		\label{fig:pod_sv2}
	\end{subfigure}
	\caption{Proper orthogonal decomposition applied to reduce the dimension of a time snapshot of the extended SEIRS model from 800 to 15 variables. The plot shows the decay of the singular values. (a) shows the singular values in a linear scale. (b) shows the singular values in a logarithm scale.}
	\label{fig:pod_sv}
\end{figure*}

In order to generate data with which to train the GAN, we performed 40 high-fidelity numerical simulations. Each simulation has two different values of $\mathcal{R}_{0\,h}$, one for people at home and another for mobile people. In the numerical simulation the whole region in Figure \ref{fig:grid_domain} is divided in a regular grid of $10\times10$, totalling $100$ cells. Although regions labelled~1 need not be modelled, solving the system on the whole domain is very efficient as a structured, regular grid can be used. Considering that each group (people at home and mobile) has four compartments in the extended SEIRS model (Susceptible,  Exposed, Infectious and Recovered), there will be eight variables for each cell in the grid per time step, which gives a total number of $8\times100=800$ variables per time step. We perform proper orthogonal decomposition in the $800$ variables, in order to work with a low dimensional space in the GAN. Figure \ref{fig:pod_sv} shows the decay of the singular values. The 15 largest singular values capture $>99.9999\%$ of the variance held in the snapshots. This was deemed sufficient, so 15 POD coefficients were retained for the NIROM. Hence the GAN is trained to generate the $15$ POD coefficients ($\bm{\alpha}^n$) and the two $\mathcal{R}_{0\,h}$ ($\bm{\mu}^n$), over a sequence of 10 time levels with a step size of two. This time length is chosen because it roughly represents a cycle (one day) in the results. 

\begin{figure*}[!htb]
	\centering
	\includegraphics[width=0.9\textwidth]{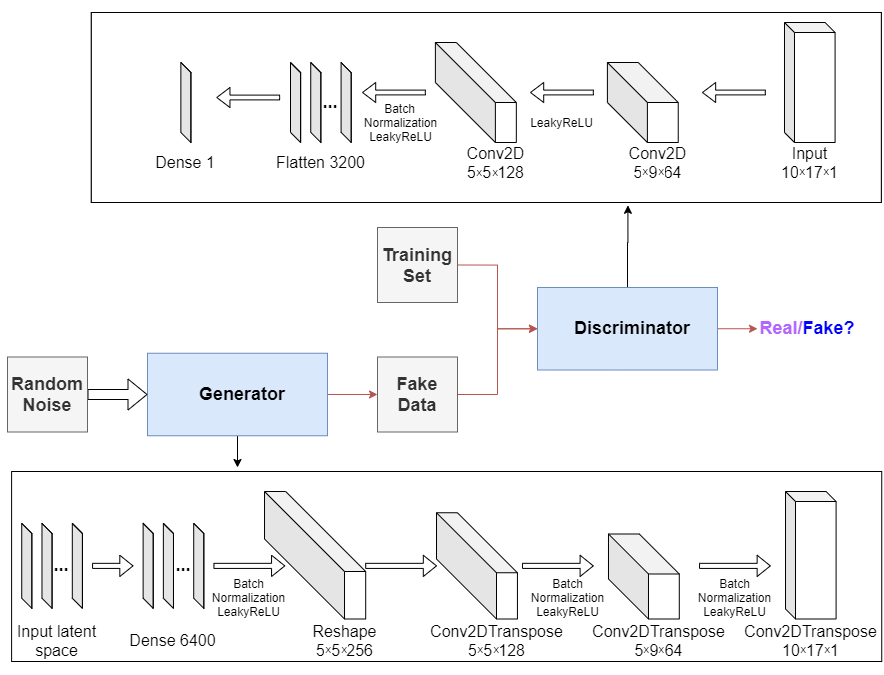}
	\caption{Generator and discriminator architectures.}
	\label{fig:gan}
\end{figure*}

The GAN architecture is based on that of the DCGAN by \cite{radford:15} and is implemented using Tensorflow \citep{tensorflow:2015}. Figure \ref{fig:gan} shows the architecture of the generator and discriminator used in this work. The generator and discriminator are trained over $5,000$ epochs, and the size of the latent vector $\mathbf{z}$ is set to 100. The 10 time levels are given to the networks as a two-dimensional array with 10 rows and 17 columns. Each row represents a time level and each column comprises the $15$ POD coefficients and the two values of $\mathcal{R}_{0\,h}$. We choose this configuration, instead of a linear representation, to take advantage of the time dependence in the two-dimensional array (``the image''). The main goal of this work is to reproduce the outputs of the high-fidelity numerical model and assimilate observed data using a GAN.  

\subsection{Predicting in space and time using the PredGAN}

In this section, we use the PredGAN (introduced in Section \ref{sec:prediction_in_time}) to make predictions of the spatial and temporal variation of the COVID-19 infection in the idealised town. All the test cases here are new or `unseen' simulations, generated from  values of $\mathcal{R}_{0\,h}$ that were not used to train the GAN. The prediction using the PredGAN is performed by starting with nine time levels from the high-fidelity numerical simulation (known initial solutions) and using the generator to predict the tenth. In the next time iteration we use eight time levels from the numerical simulation and the last prediction to predict a new point. Then we repeat this process until the last time step. It is worth mentioning that after nine time iterations, the PredGAN works only with data from the predictions. Data from the high-fidelity numerical simulation is used only for the first nine time levels as an initial condition.   

\begin{figure*}[!h]
	\centering
	\begin{subfigure}[t]{0.9\textwidth}
		\centering
		\includegraphics[width=\textwidth]{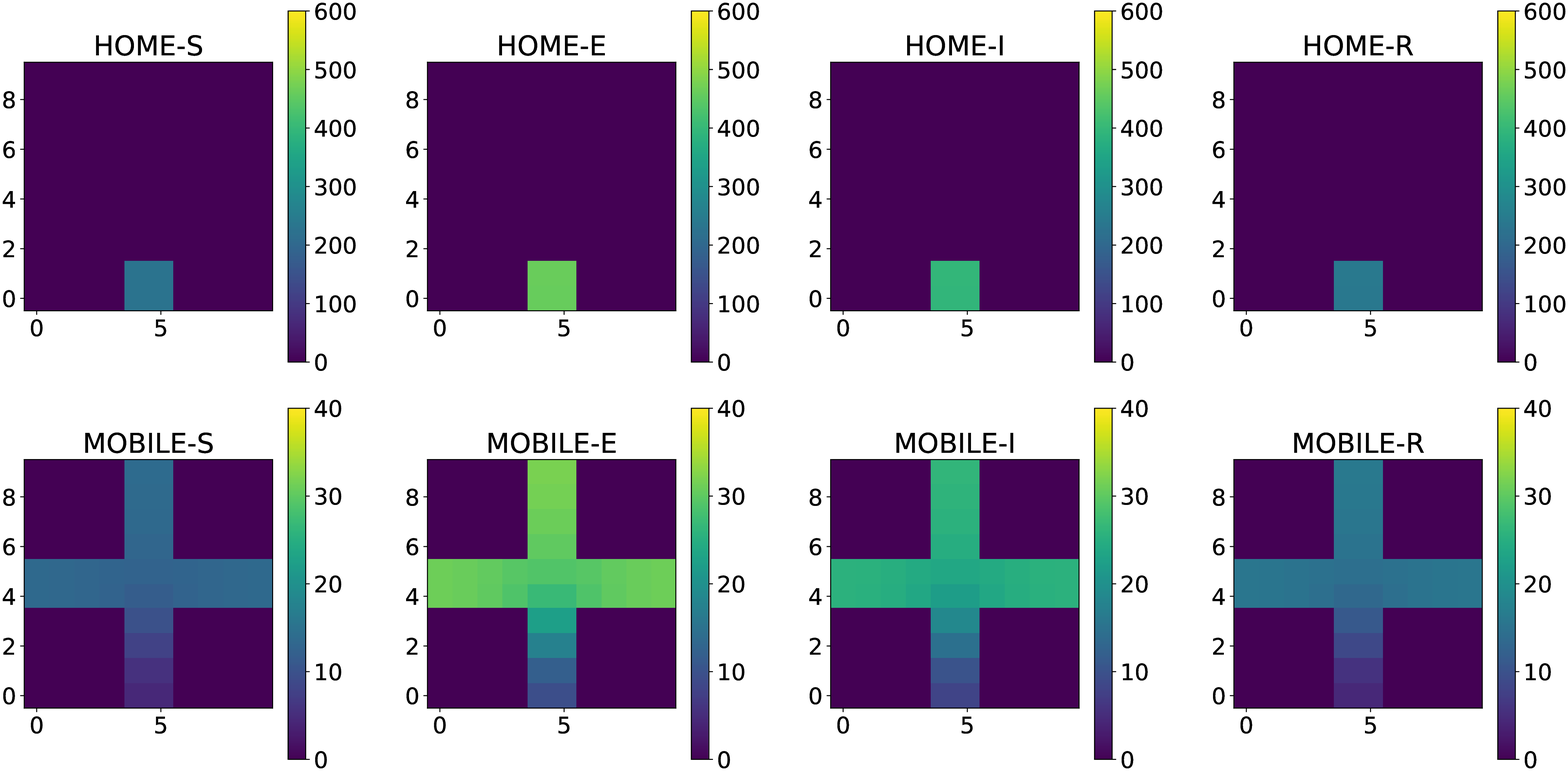}
		\caption{Prediction.}
		\label{fig:meshrun2}
	\end{subfigure}
	\begin{subfigure}[t]{0.9\textwidth}
		\centering
		\includegraphics[width=\textwidth]{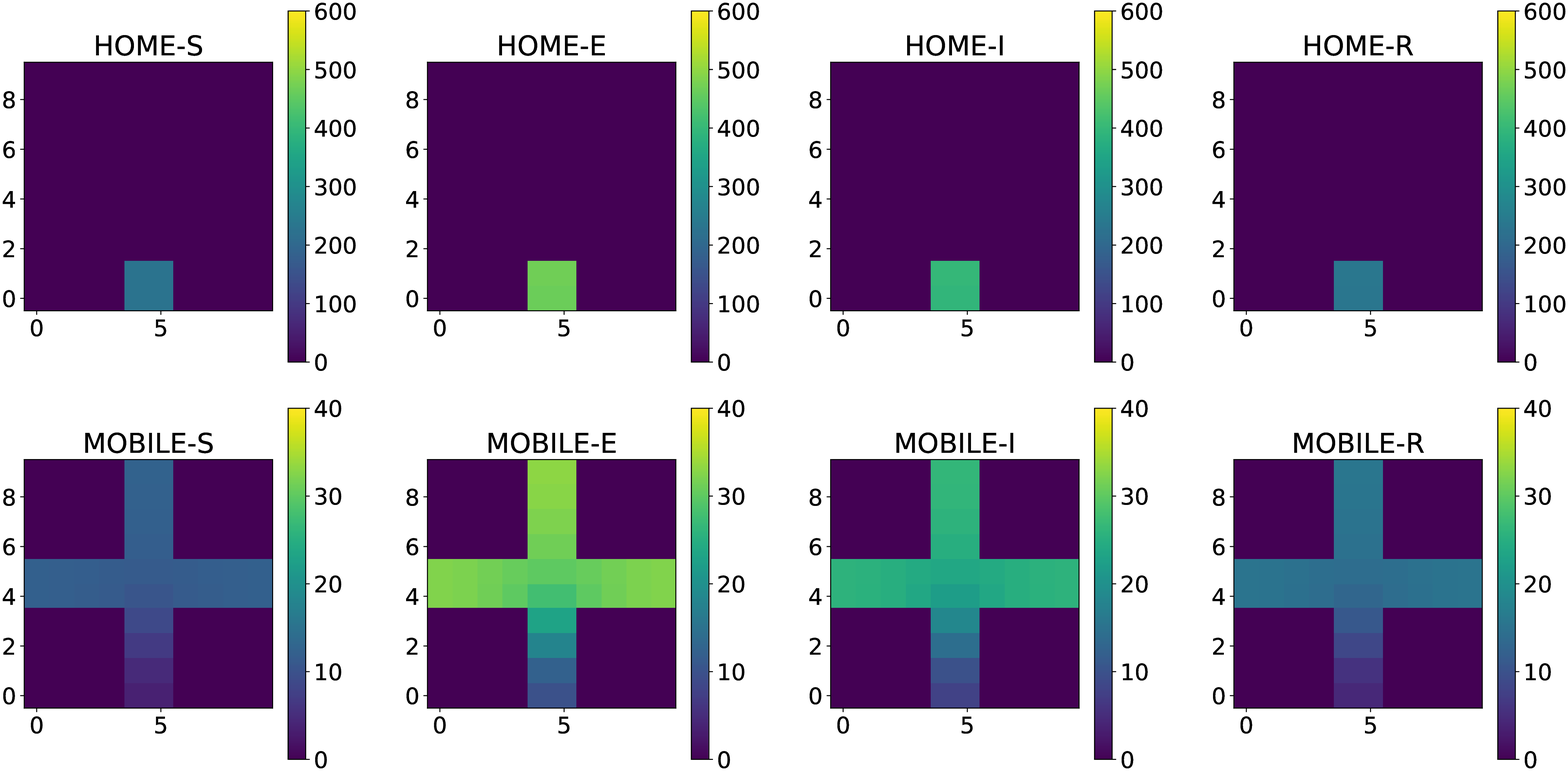}
		\caption{Ground truth.}
		\label{fig:meshrun2real}
	\end{subfigure}
	\begin{subfigure}[t]{0.9\textwidth}
		\centering
		\includegraphics[width=\textwidth]{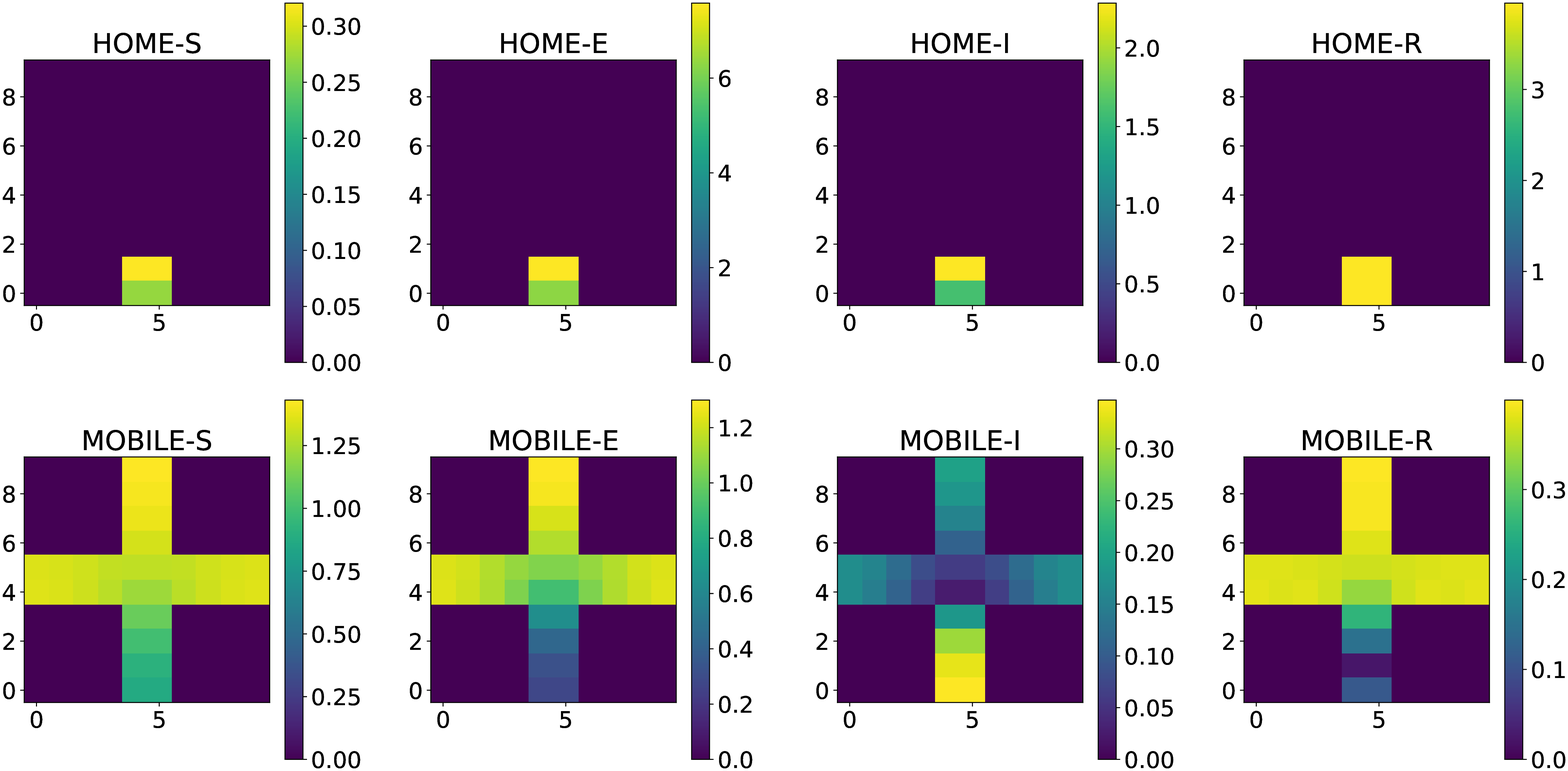}
		\caption{Absolute Difference.}
		\label{fig:meshrun2diff}
	\end{subfigure}
	\caption{PredGAN applied to predict one time level ($\mathcal{R}_{0\,1}=7.7$ and $\mathcal{R}_{0\,2}=17.4$). (a) shows the prediction of the number of people across the domain. (b) shows the number of people across the domain calculated by the high-fidelity numerical simulation. (c) shows the difference between the PredGAN and the high-fidelity numerical simulation.}
	\label{fig:meshrun2all}
\end{figure*}

The first result we present here is the prediction of one time level for the values $\mathcal{R}_{0\,1}=7.7$ and $\mathcal{R}_{0\,2}=17.4$. The first nine time levels (initial condition) were taken from the high-fidelity numerical simulation after 21 days from the start of the infection. Figure \ref{fig:meshrun2all} shows the spatial variation of the number of people in each group and compartment throughout the domain. Figure \ref{fig:meshrun2} shows the prediction of the PredGAN, Figure \ref{fig:meshrun2real} shows the actual result from the numerical simulation, and Figure \ref{fig:meshrun2diff} the absolute difference between them. All the quantities represent the number of people in a cell of the domain. The mean absolute error between the ground truth and the prediction is $0.19$ and the relative mean absolute error is $8.9\times10^{-3}$. These results show that the PredGAN was able to predict accurately the evolution of the extended SEIRS model in space and time, at least for one time iteration.  

\begin{figure*}[!h]
	\centering
	\includegraphics[width=1.0\textwidth]{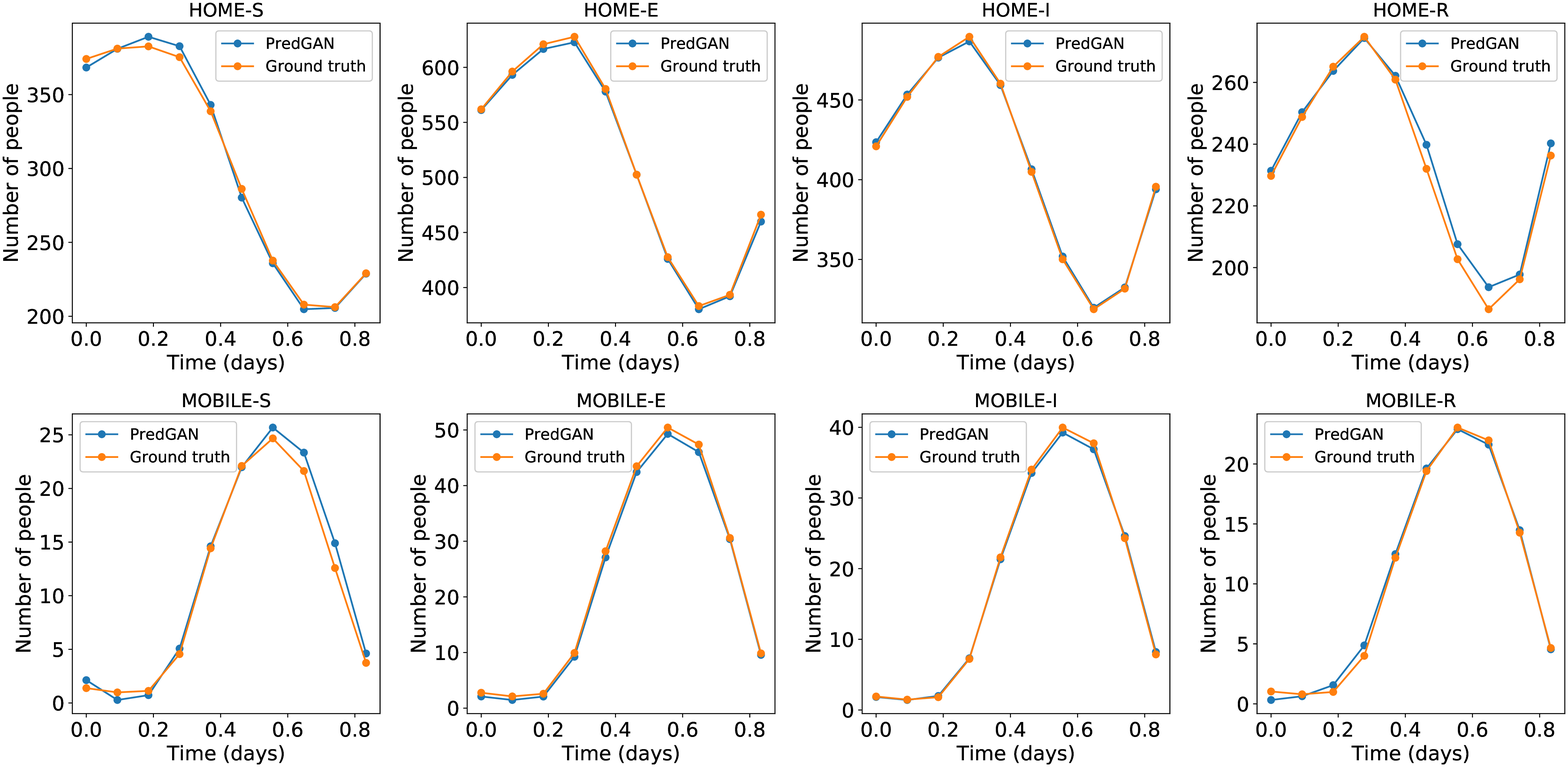}
	\caption{Prediction of one time level of the spatial variation COVID-19 infection ($\mathcal{R}_{0\,1}=7.7$ and $\mathcal{R}_{0\,2}=17.4$). The results show the time variation in one cell of the grid (bottom-left corner of region 2). The first nine points are used to start the PredGAN and the last one is the actual prediction.}
	\label{fig:run2_1ts}
\end{figure*}

Figure \ref{fig:run2_1ts} shows the same set of results as in Figure~\ref{fig:meshrun2all}, except now, we focus on the prediction at one point in space or one cell in the domain (bottom-left corner of region 2 in Figure \ref{fig:grid_domain}). Each plot corresponds to the variation of the number of people in each group and compartment over time. The first nine time levels are used in the optimisation process of the PredGAN and the tenth time level is actually the prediction of the unknown solution. When using PredGAN, solutions for the POD coefficients at all 10 time levels are obtained, and all 10 time levels are shown here for illustration. As explained in Section~\ref{sec:prediction_in_time}, PredGAN minimises the difference between the nine known values and its predictions at these time levels. Once converged, PredGAN's prediction for the tenth time level is accepted. There will be small differences between the known values and predicted values for the first nine time levels, which are shown in Figure~\ref{fig:run2_1ts}, but these are ignored in future calculations as only the tenth time level is added to the known solutions. Comparable results regarding the error in the prediction were seen at other points in the domain, therefore we do not present them here. It can be noticed from Figures \ref{fig:meshrun2all} and \ref{fig:run2_1ts} that the PredGAN can reasonably predict the outcomes of the high-fidelity numerical model for simulations that are not in training set of the GAN.   

\begin{figure*}[!h]
	\centering
	\includegraphics[width=1.0\textwidth]{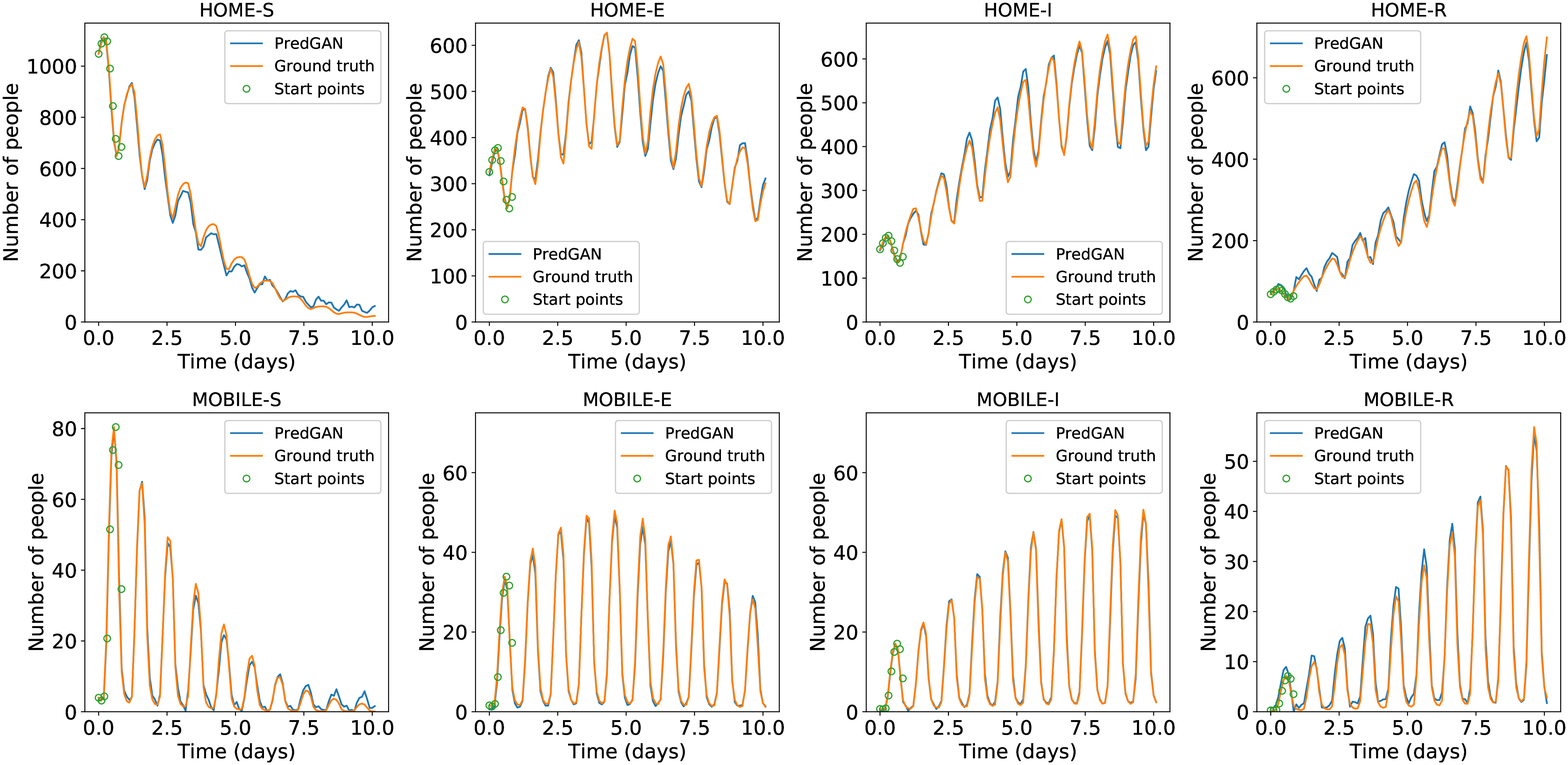}
	\caption{Prediction of multiple time levels of the spread of the COVID-19 infection ($\mathcal{R}_{0\,1}=7.7$ and $\mathcal{R}_{0\,2}=17.4$). The results show the time variation in one cell of the grid (bottom-left corner of region 2). The first nine points (indicated by green circles) are used to start the PredGAN and all the others are predictions.}
	\label{fig:run2}
\end{figure*}

\begin{figure*}[!h]
	\centering
	\begin{subfigure}[t]{1.0\textwidth}
		\centering
		\includegraphics[width=\textwidth]{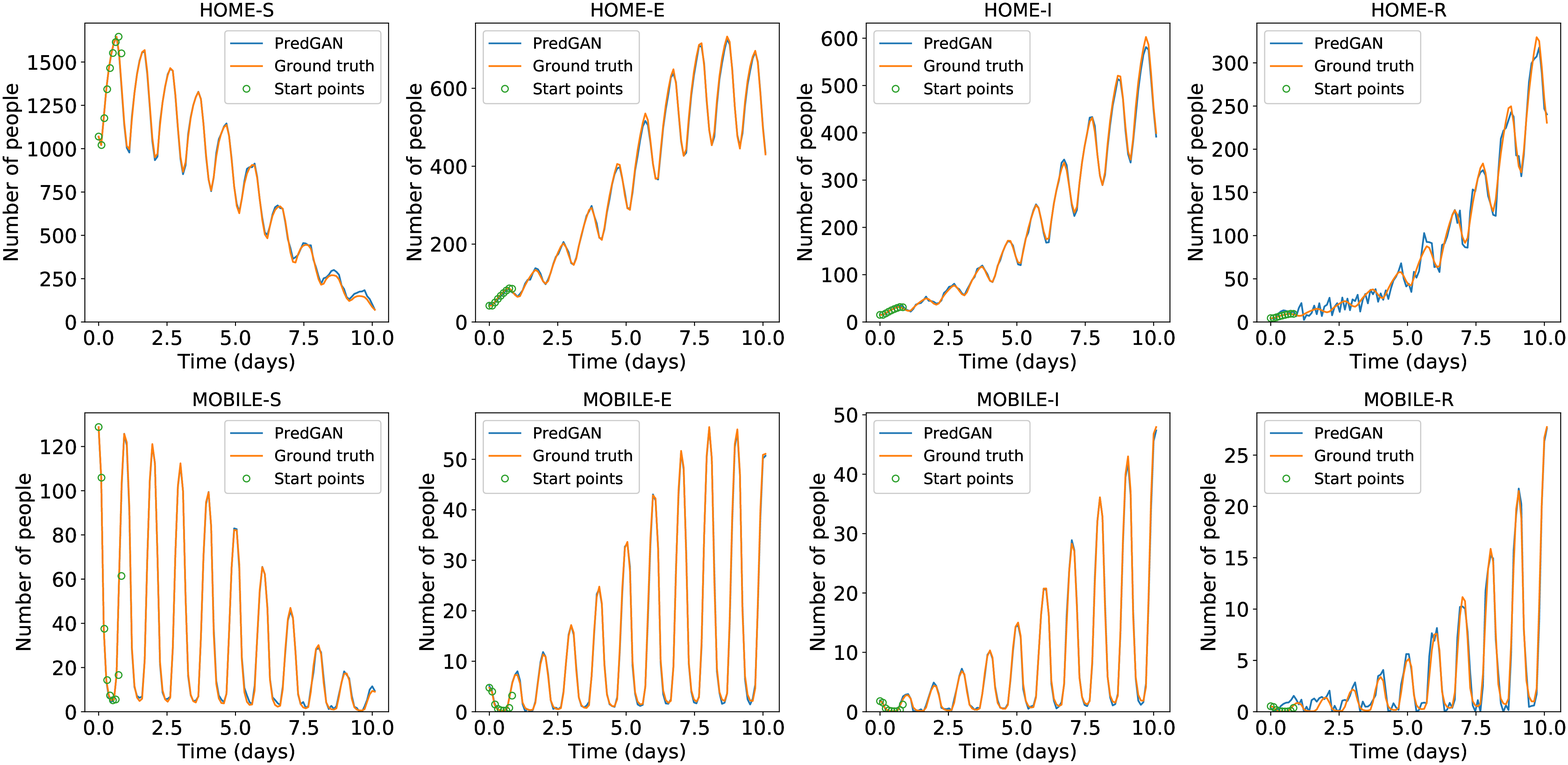}
		\caption{$\mathcal{R}_{0\,1}$ = 14.7, $\mathcal{R}_{0\,2}$ = 9.6.}
		\label{fig:run16}
	\end{subfigure}
	\begin{subfigure}[t]{1.0\textwidth}
		\centering
		\includegraphics[width=\textwidth]{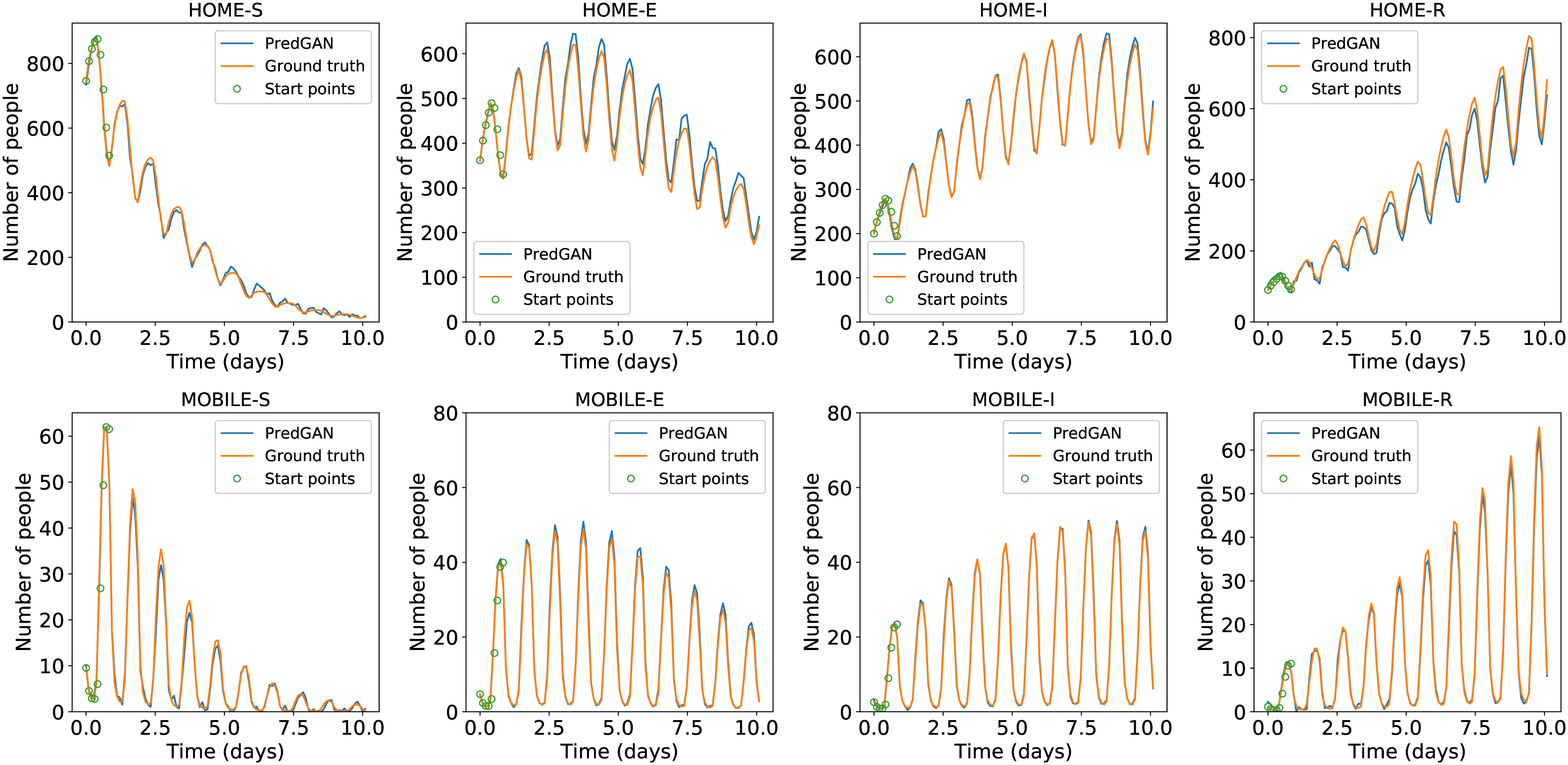}
		\caption{$\mathcal{R}_{0\,1}$ = 9.5, $\mathcal{R}_{0\,2}$ = 10.1.}
		\label{fig:run17}
	\end{subfigure}
	\caption{Prediction of multiple time levels of the spatial variation COVID-19 infection for different basic reproduction numbers. The results show the time variation in one cell of the grid (bottom-left corner of region 2). The first nine points are used to start the PredGAN and all the others are predictions.}
	\label{fig:run16and17}
\end{figure*}

For predicting further in time, we again start with nine time levels from the high-fidelity numerical simulation and we use the generator to predict the tenth. The next iterations we use the last predictions as known values and we repeat this process until the end of the simulation. Figure \ref{fig:run2} shows the result of the prediction in one cell of the grid (bottom-left corner of region 2 in Figure \ref{fig:grid_domain}). Each cycle corresponds to a period of one day, when mobile people leave their homes during the day and return at night. After the first nine time iterations the PredGAN does not see any data from the high-fidelity numerical simulation, and relies completely on the predictions from PredGAN. Data from the high-fidelity numerical simulation is only required as an initial condition. Figure  \ref{fig:run16and17} also shows the prediction of the PredGAN, although this time using two more different sets of values for $\mathcal{R}_{0\,h}$ and starting at different times of the epidemic dynamics. The results presented in Figures \ref{fig:run2} and \ref{fig:run16and17} indicate that the prediction of multiple time levels was very successful. For all compartments and groups, the prediction using the PredGAN is almost indistinguishable from the ground truth or high-fidelity numerical simulation. Hence the PredGAN can be used as a surrogate model of high-fidelity numerical simulations varying in space and time.   

%\clearpage
\subsection{Data Assimilation using the DA-PredGAN}

In this section, we apply the DA-PredGAN (introduced in Section \ref{sec:data_assimilation}) to assimilate observed data to the spatial variation of COVID-19 over time. The data assimilation using the DA-PredGAN works similarly to the PredGAN, apart from adding an observed data mismatch term in the functional (Eqs.~\eqref{eq:loss_da_forward} and~\eqref{eq:loss_da_backward}), not knowing the model parameters $\mathcal{R}_{0,h}$ a priori, and from working forwards and backwards in time. We generate observed data from a high-fidelity numerical simulation that was not included in the training set of the GAN. To that end, we use $\mathcal{R}_{0\,1}=7.7$, $\mathcal{R}_{0\,2}=17.4$, and also add $5\%$ noise to the chosen data. Considering the domain in Figure \ref{fig:grid_domain}, we choose to have observed data collected at the bottom-left corner of regions 2, 3, 4, 5 and 6. In other words, the observed data is available at five points in domain, one in the middle and one at each end of the cross shaped region. The $\mathcal{R}_{0\,h}$ are not used as observed data, although we compare it with the true values used to generate the high-fidelity numerical simulation. To start the DA-PredGAN, we perform one forward march without the observed data term in the functional (as described in Section \ref{sec:dattime}). The starting points chosen for this march are from a numerical simulation with $\mathcal{R}_{0\,1}=6.5$ and $\mathcal{R}_{0\,2}=5.7$. 

\begin{figure*}[!htb]
	\centering
	\begin{subfigure}[t]{0.48\textwidth}
		\centering
		\includegraphics[width=\textwidth]{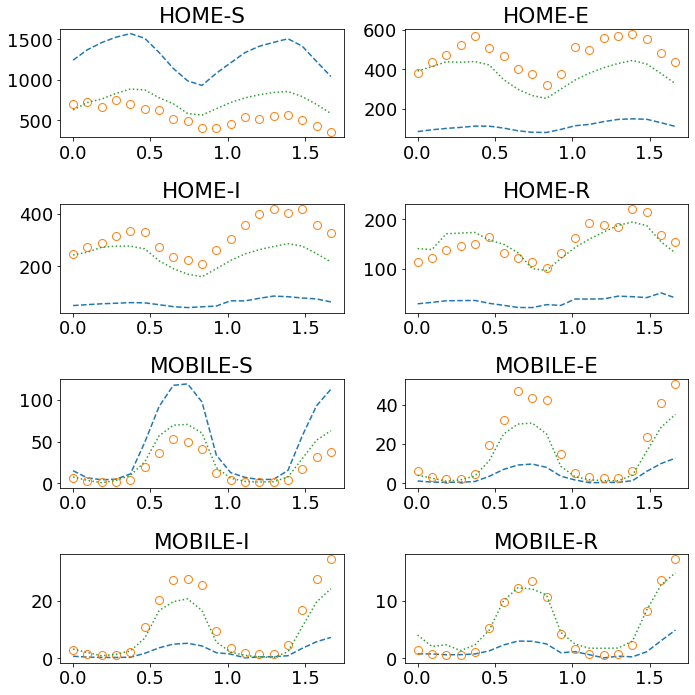}
		\caption{Iteration 1.}
		\label{fig:run2iter1}
	\end{subfigure} \;
	\begin{subfigure}[t]{0.48\textwidth}
		\centering
		\includegraphics[width=\textwidth]{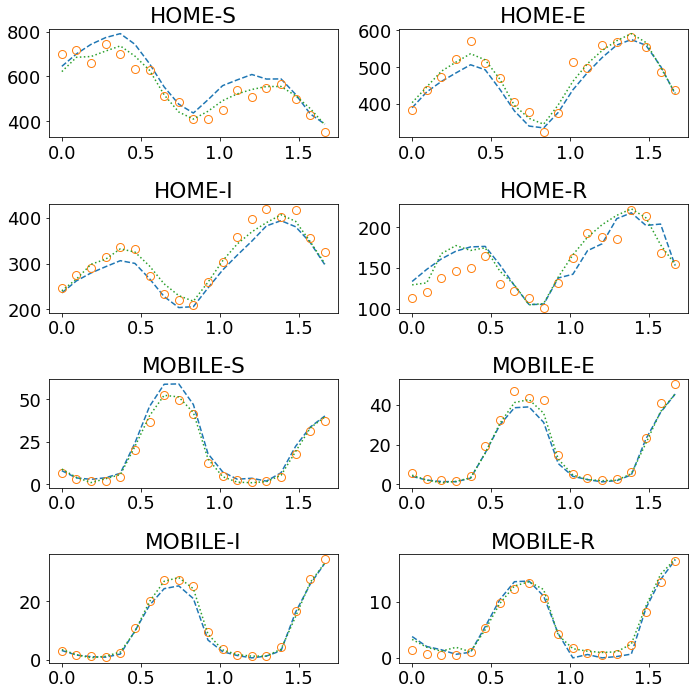}
		\caption{Iteration 2.}
		\label{fig:run2iter2}
	\end{subfigure}
	\begin{subfigure}[t]{0.48\textwidth}
		\centering
		\includegraphics[width=\textwidth]{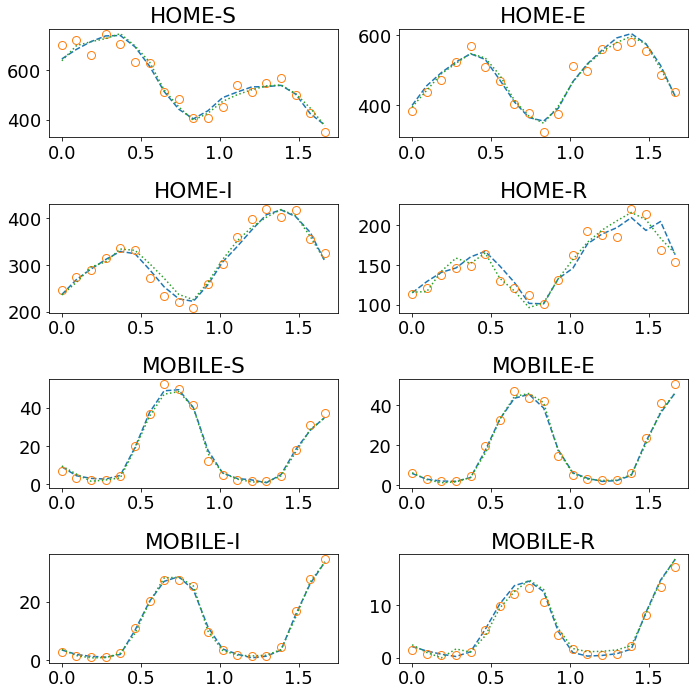}
		\caption{Iteration 10.}
		\label{fig:run2iter10}
	\end{subfigure} \;
	\begin{subfigure}[t]{0.48\textwidth}
		\centering
		\includegraphics[width=\textwidth]{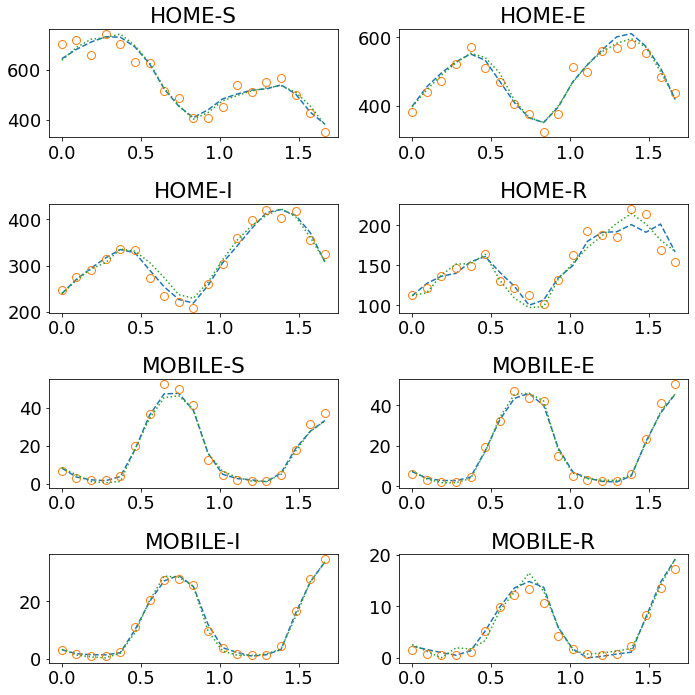}
		\caption{Iteration 30.}
		\label{fig:run2iter30}
	\end{subfigure}	
	\caption{ Forward-backward iterations 1 2, 10 and 30 of the DA-PredGAN applied to spatial variation of COVID-19 infection. The results show the time variation of groups and compartments in one cell of the grid (bottom-left corner of region 2). The horizontal axes are the time in days and the vertical axes are the number of people. The orange circles represent the observed data, the dashed blue line the forward march, and the dotted green line the backward march. }
	\label{fig:run2iters}
\end{figure*}

\begin{figure}[!htb]
    \centering
    \begin{subfigure}[t]{0.6\textwidth}
    	\centering
    	\includegraphics[width=\textwidth]{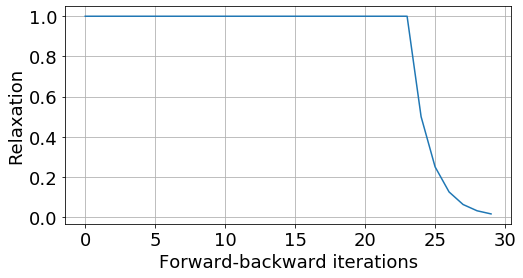}
    	\caption{}
    	\label{fig:run2relax}
    \end{subfigure}
    \begin{subfigure}[t]{0.6\textwidth}
    	\centering
    	\includegraphics[width=\textwidth]{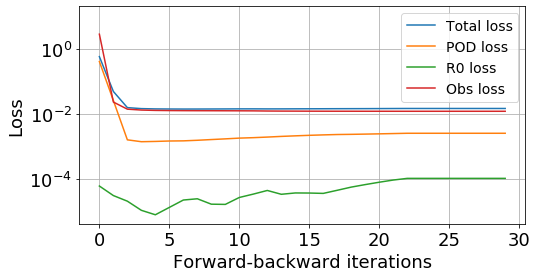}
    	\caption{}
    	\label{fig:run2loss}
    \end{subfigure}
	\caption{Evolution of the data assimilation process. (a) relaxation factor $r^j$. The convergence criteria is reached when $r^j<0.01$. (b) loss functions of the DA-PredGAN. The curves represent the average total loss and the average values of each term in the Eqs.~\eqref{eq:loss_da_forward} and \eqref{eq:loss_da_backward}.}
	\label{fig:run2relaxloss}
\end{figure}

\begin{figure*}[!htb]
    \centering
	\begin{subfigure}[t]{0.32\textwidth}
		\centering
		\includegraphics[width=\textwidth]{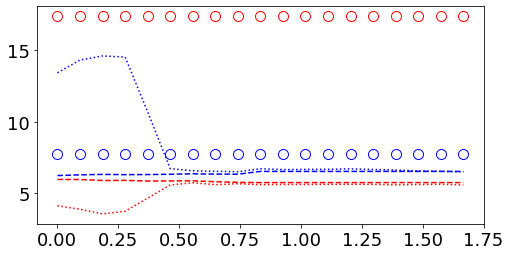}
		\caption{Iteration 1.}
		\label{fig:R0run2iter1}
	\end{subfigure}
	\begin{subfigure}[t]{0.32\textwidth}
		\centering
		\includegraphics[width=\textwidth]{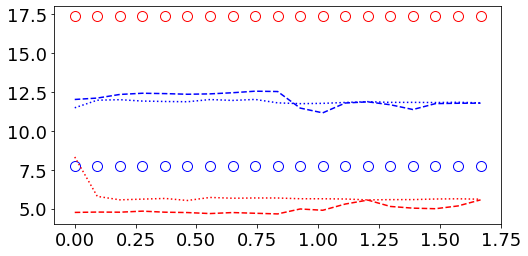}
		\caption{Iteration 2.}
		\label{fig:R0run2iter2}
	\end{subfigure}
	\begin{subfigure}[t]{0.32\textwidth}
		\centering
		\includegraphics[width=\textwidth]{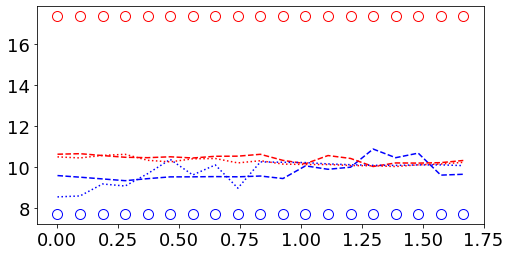}
		\caption{Iteration 4.}
		\label{fig:R0run2iter4}
	\end{subfigure}
	\begin{subfigure}[t]{0.32\textwidth}
		\centering
		\includegraphics[width=\textwidth]{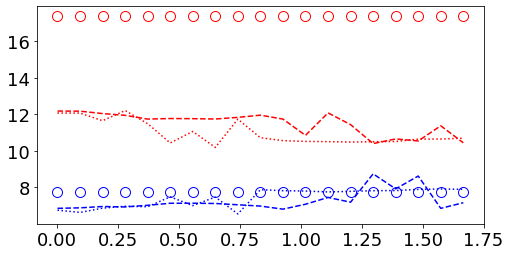}
		\caption{Iteration 8.}
		\label{fig:R0run2iter8}
	\end{subfigure}
	\begin{subfigure}[t]{0.32\textwidth}
		\centering
		\includegraphics[width=\textwidth]{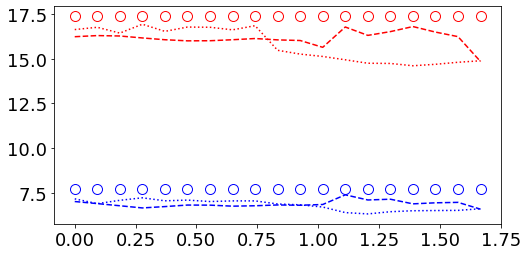}
		\caption{Iteration 16.}
		\label{fig:R0run2iter16}
	\end{subfigure}
	\begin{subfigure}[t]{0.32\textwidth}
		\centering
		\includegraphics[width=\textwidth]{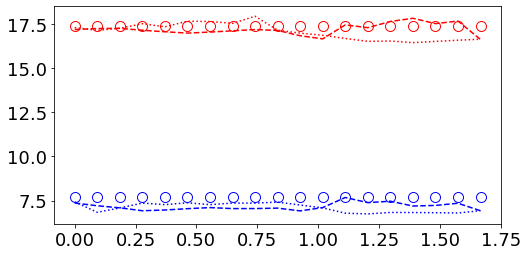}
		\caption{Iteration 30.}
		\label{fig:R0run2iter30}
	\end{subfigure}
	\caption{Evolution of the basic reproduction number ($\mathcal{R}_{0\,h}$) during the assimilation process of the DA-PredGAN. Each plot represents a forward-backward iteration. The horizontal axes are the time in days and the vertical axes are the $\mathcal{R}_{0\,h}$. The circles represent the true value, the dashed lines the forward marches, and the dotted lines the backward marches. Blue represents the home group ($\mathcal{R}_{0\,1}$) and red the mobile group ($\mathcal{R}_{0\,2}$).}
	\label{fig:R0run2iters}
\end{figure*}

\begin{figure*}[!htb]
	\centering
	\includegraphics[width=1.0\textwidth]{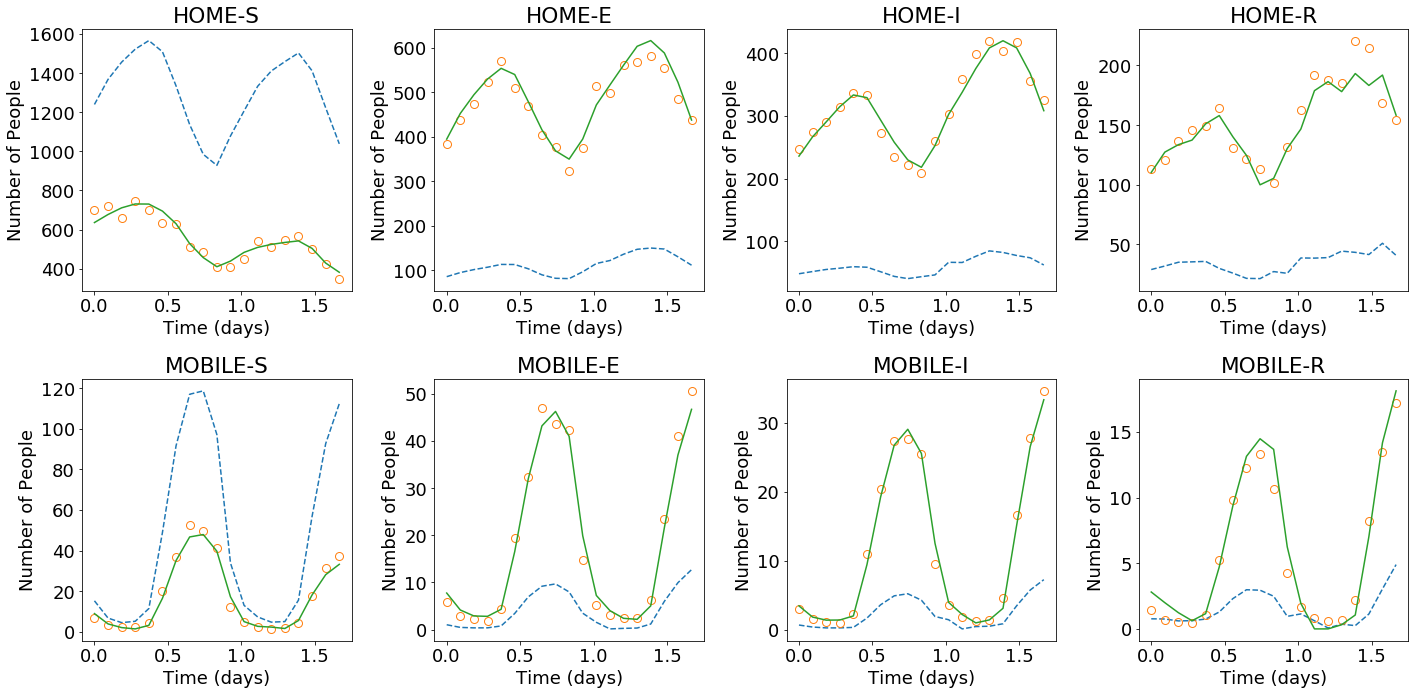}
	\caption{Initial and final results of the DA-PredGAN applied to spatial variation of COVID-19 infection. The results show the time variation of groups and compartments in one cell of the grid (bottom-left corner of region 2). The orange circles represent the observed data, the dashed blue line is the first forward march, and the solid green line is the final result (last forward march).}
	\label{fig:run2inivsfinal}
\end{figure*}

\begin{figure}[!htb]
	\centering
	\includegraphics[width=0.7\textwidth]{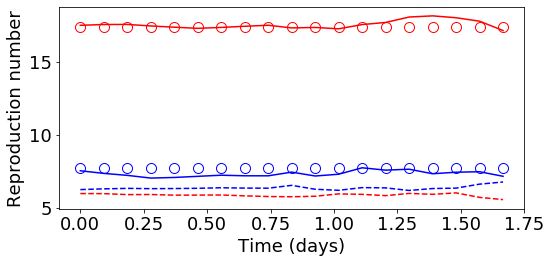}
	\caption{Initial and final values of the basic reproduction number ($\mathcal{R}_{0\,h}$) during the assimilation process of the DA-PredGAN. The circles represent the true value, the dashed lines the first forward march, and the solid lines the final result (last forward march). Blue represents the home group ($\mathcal{R}_{0\,1}$) and red the mobile group ($\mathcal{R}_{0\,2}$).}
	\label{fig:R0run2inivsfinal}
\end{figure}

Figure \ref{fig:run2iters} shows the evolution of the forward-backward iterations of the data assimilation process using the DA-PredGAN. The results show the time variation of groups and compartments in one cell of the grid (bottom-left corner of regions 2). It can be seen from this figure that in just a few forward-backward iterations the DA-PredGAN is able to match the data. Although we run the simulation until the convergent criteria was reached (see Figure \ref{fig:run2relax}), after iteration 2, only small improvements in the observed data mismatch can be noticed. This is also shown in Figure \ref{fig:run2loss}, along with the average total loss and the other average loss terms in the functional (Eqs.~\eqref{eq:loss_da_forward} and~\eqref{eq:loss_da_backward}). The evolution of the $\mathcal{R}_{0\,h}$ for the same data assimilation is presented in Figure \ref{fig:R0run2iters}. The result shows that as long as the data mismatch is minimised, the model parameters $\mathcal{R}_{0\,h}$ approach the true values used to generate the synthetic observed data. We also present in Figures \ref{fig:run2inivsfinal} and \ref{fig:R0run2inivsfinal} a comparison between the first and the last forward iterations. These figures show that even with the initial guess far from the observed data the method was able to match the measurements and produce model parameters $\mathcal{R}_{0\,h}$ near the true value.                

\begin{figure*}[!htb]
	\centering
	\includegraphics[width=1.0\textwidth]{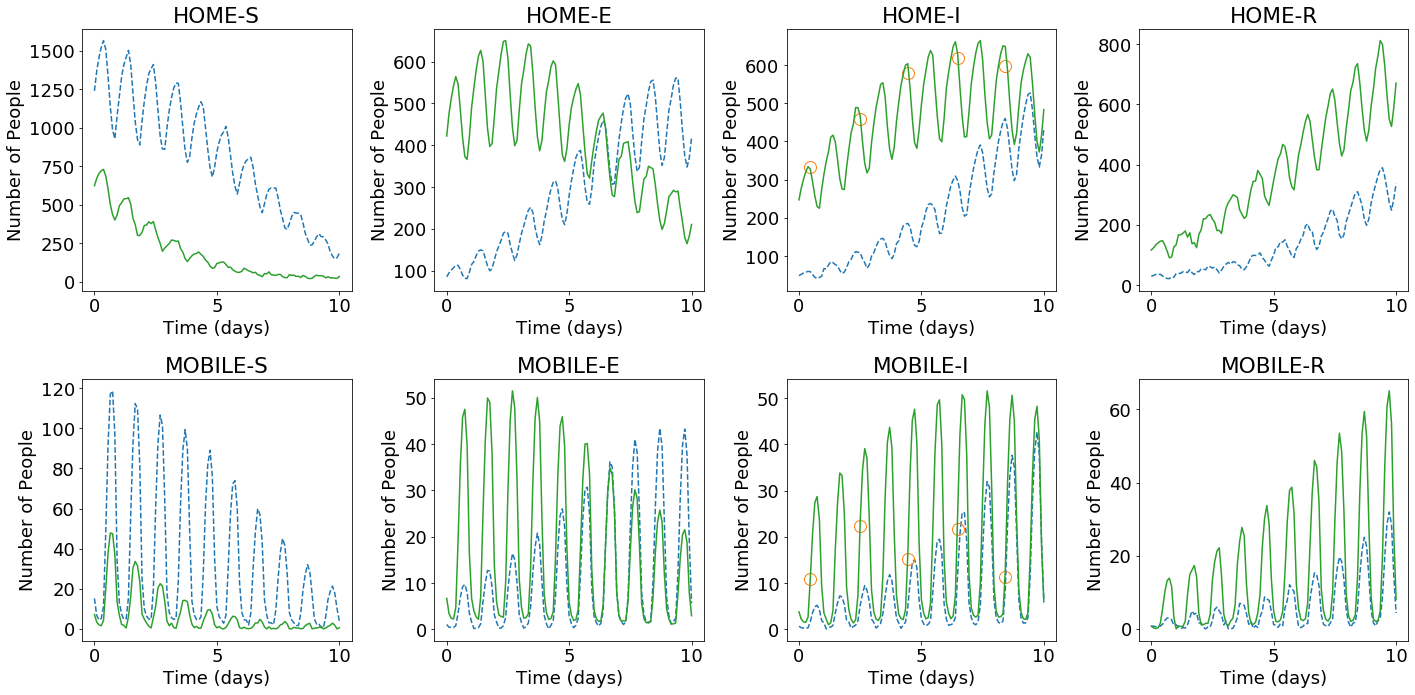}
	\caption{Initial and final results of the DA-PredGAN applied to spatial variation of COVID-19 infection for the more realistic case. The results show the time variation of groups and compartments in one cell of the grid (bottom-left corner of region 2). The orange circles represent the observed data, the dashed blue line is the first forward march, and the solid green line is the final result (last forward march).}
	\label{fig:run2inivsfinalmult}
\end{figure*}

\begin{figure}[!htb]
	\centering
	\includegraphics[width=0.7\textwidth]{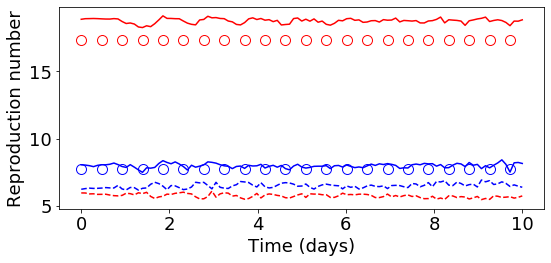}
	\caption{Initial and final values of the basic reproduction number ($\mathcal{R}_{0\,h}$) during the assimilation process of the DA-PredGAN for the more realistic case. The circles represent the true value, the dashed lines the first forward march, and the solid lines the final result (last forward march). Blue represents the home group ($\mathcal{R}_{0\,1}$) and red the mobile group ($\mathcal{R}_{0\,2}$).}
	\label{fig:R0run2inivsfinalmult}
\end{figure}

\begin{figure}[!htb]
    \centering
    \begin{subfigure}[t]{0.6\textwidth}
    	\centering
    	\includegraphics[width=\textwidth]{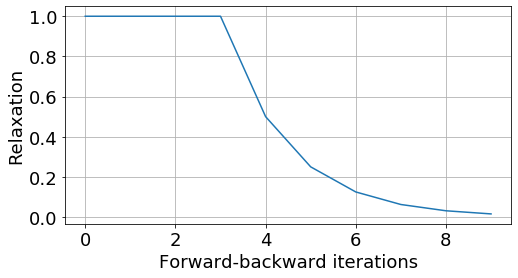}
    	\caption{}
    	\label{fig:run2relaxmult}
    \end{subfigure}
    \begin{subfigure}[t]{0.6\textwidth}
    	\centering
    	\includegraphics[width=\textwidth]{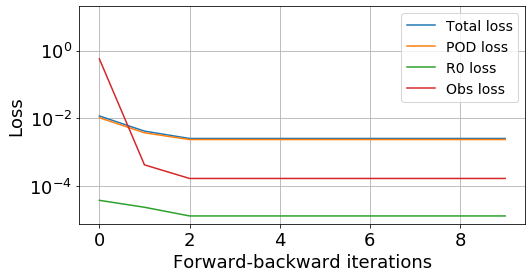}
    	\caption{}
    	\label{fig:run2lossmult}
    \end{subfigure}
	\caption{Evolution of the data assimilation process for the more realistic case. (a) relaxation factor $r^j$. The convergence criteria is reached when $r^j<0.01$. (b) loss functions of the DA-PredGAN. The curves represent the average total loss and the average values of each term in the Eqs.~\eqref{eq:loss_da_forward} and \eqref{eq:loss_da_backward}.}
	\label{fig:run2relaxlossmult}
\end{figure}

\begin{figure*}[!htb]
    \centering
	\begin{subfigure}[t]{1.0\textwidth}
		\centering
		\includegraphics[width=\textwidth]{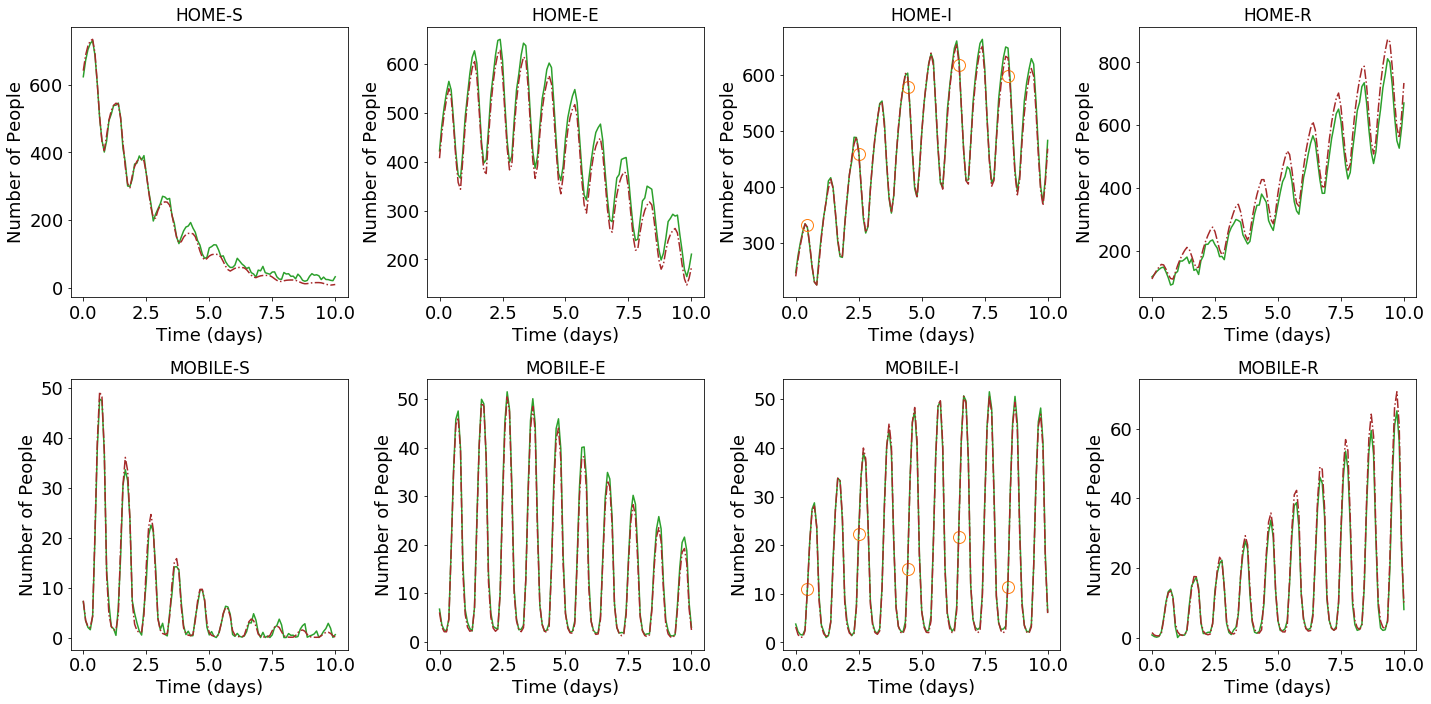}
		\caption{}
		\label{fig:run2finalvstruth1}
	\end{subfigure}
	\begin{subfigure}[t]{1.0\textwidth}
		\centering
		\includegraphics[width=\textwidth]{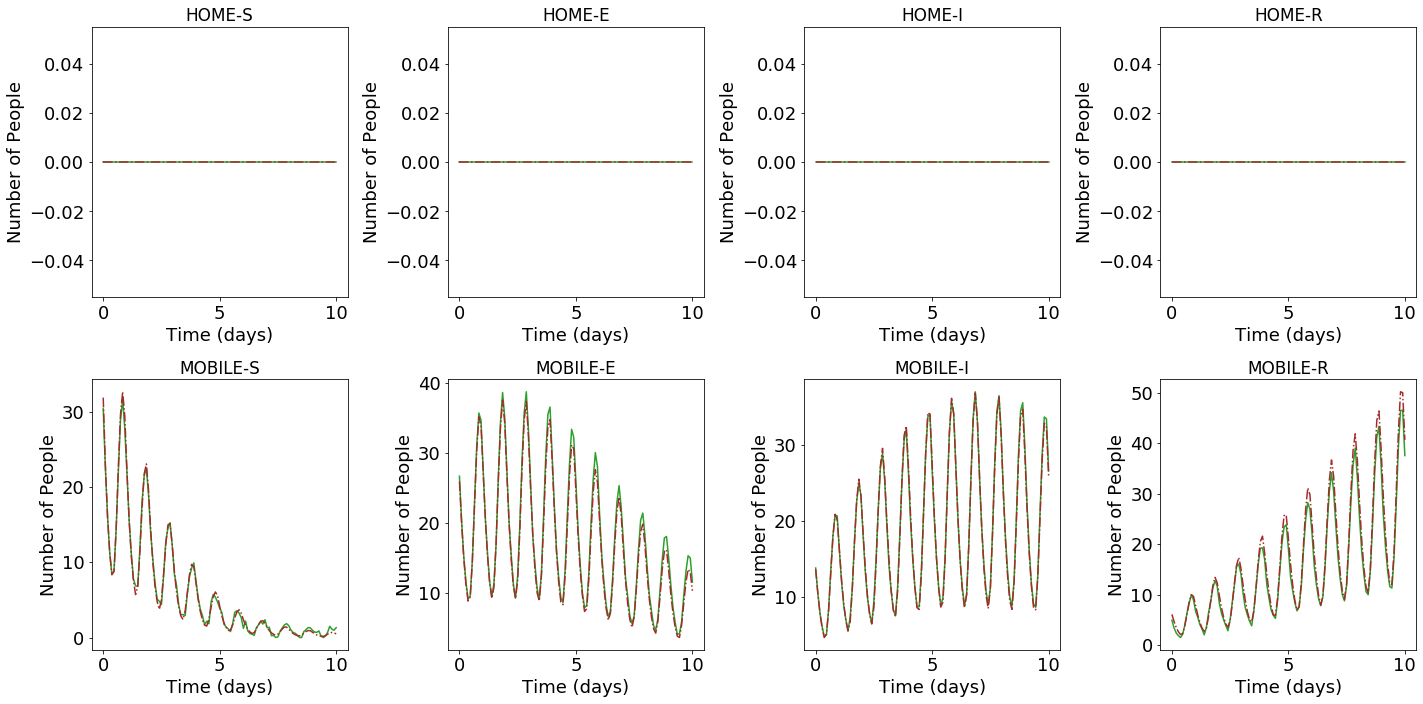}
		\caption{}
		\label{fig:run2finalvstruth2}
	\end{subfigure}
	\caption{Final results of the DA-PredGAN compared with the high-fidelity numerical simulation used to generate the observed data. The orange circles represent the observed data, the dashed brown line is the high-fidelity numerical simulation, and the solid green line is the result of the DA-PredGAN (last forward march). (a) shows the time variation of groups and compartments in one cell of the grid (bottom-left corner of region 2) with observed data. (b) shows the time variation of groups and compartments in one cell of the grid (top-right corner of region 10) without observed data.}
	\label{fig:run2finalvstruth}
\end{figure*}

In order to test the DA-PredGAN in a more realistic case, we consider that observed data is only available every two days, and we measure only infectious people. Figures \ref{fig:run2inivsfinalmult} and \ref{fig:R0run2inivsfinalmult} show the first and last forward marches of the data assimilation process. We observe that the method proposed here was able to effectively match the observed data and to produce model parameters $\mathcal{R}_{0\,h}$ with similar values as the ones used to generate the synthetic data. It is worth noticing that the data assimilation is an inverse and usually ill-posed problem, thus other values of $\mathcal{R}_{0\,h}$ could have also matched the observed data, within some tolerance. Figure \ref{fig:run2relaxlossmult} shows the relaxation factor and the loss terms of the DA-PredGAN over the forward-backward iterations. These results demonstrate the efficiency of the DA-PredGAN, since it was capable of matching the observed data in only few iterations even starting far from the measurements. We also present in Figure \ref{fig:run2finalvstruth} a comparison between the DA-PredGAN results and the high-fidelity numerical simulation used to generate the observed data. Figure \ref{fig:run2finalvstruth1} shows the evolution of the number of people in each group and compartment for a point in space where observed data was collected. Figure \ref{fig:run2finalvstruth2} shows the same plots, but for a point in space without observed data. Although we would not expect that the results of the data assimilation will reproduce the ``true'' simulation, since it is a ill-posed inverse problem, we observe from these figures that the DA-PredGAN was able to generate coherent results that resemble the dynamics of the ground truth, even at points where observed data was not collected.    

\section{Discussion}
\label{sec:disc}

Despite one of the original purposes of generative adversarial networks (GANs), to be able to generate realistic-looking images, this paper demonstrates that GANs can also be used to perform spatio-temporal prediction (PredGAN algorithm) and data assimilation (DA-PredGAN algorithm). The GAN was chosen here because incredible results have been achieved with this network, clearly outperforming other methods in many applications. However, other generative models could also fit into the PredGAN and DA-PredGAN algorithms. We also remark that, although here the proposed methods are set within a non-intrusive reduced-order model (NIROM) framework, these algorithms could be based directly on the high-dimensional system. The NIROM was used to reduce the number of degrees of freedom which makes training the GANs more manageable.

Focusing on the DA-PredGAN, it has the following advantages and disadvantages relative to other data assimilation algorithms. The advantages are that the DA-PredGAN has potentially more rapid convergence properties, as even within a few forward-backward iterations the method was able to match the observed data and update the model parameters. Furthermore, no additional simulation of the high-fidelity numerical model is needed to assimilate data using the DA-PredGAN. Another advantage is the use of the inherent adjoint capabilities of neural networks to calculate the gradients. The error in the loss functions is back-propagated through the network using the available machine learning codes e.g.~Tensorflow, PyTorch. The primary disadvantage is the need to tuning the weighting terms $\zeta_{obs}$ and $\zeta_\mu$ in the loss functions. If not adjusted the method may change the solution variables prematurely within a forward-backward iteration, or conversely, just make very small changes to them. To tackle this problem, we have proposed some values for the weighting terms in Section \ref{sec:weighting_terms}. These values have worked well for all the cases we have run.

\section{Conclusion}
\label{sec:conclusion}

In this work, we proposed a generative adversarial network that is able to make predictions in space and time (PredGAN), and we set this within a reduced-order model framework for efficiency. The aim of the PredGAN is to be a surrogate model of the high-fidelity numerical simulation. Furthermore, we extended the forecast using generative adversarial networks to assimilate observed data (DA-PredGAN) without any additional simulations of the high-fidelity numerical model. We applied these approaches to an extended SEIRS model to predict the spread of COVID-19 over space and time. The results show that the surrogate model is able to accurately reproduce the numerical simulation for different model inputs. We also demonstrate the efficiency of the DA-PredGAN in assimilating observed data and determining the corresponding model parameters. The proposed methods may have important implications for a huge class of physical simulation problems, for developing accurate surrogate models and efficiently assimilating measurements.         

\clearpage
\section*{Acknowledgements}
 This work is supported by the following EPSRC grants: RELIANT, Risk EvaLuatIon fAst iNtelligent Tool for COVID19 (EP/V036777/1); MUFFINS, MUltiphase Flow-induced Fluid-flexible structure InteractioN in Subsea applications (EP/P033180/1); the PREMIERE programme grant (EP/T000414/1); INHALE, Health assessment across biological length scales (EP/T003189/1); and MAGIC, Managing Air for Green Inner Cities (EP/N010221/1). This work has been undertaken, in part, as a contribution to `Rapid Assistance in Modelling the Pandemic' (RAMP), initiated by the Royal Society. In particular, we would like to acknowledge the useful discussion had within the Environmental and Aerosol Transmission group of RAMP, coordinated by Profs Paul Linden and Christopher Pain. The first author also acknowledges the financial support from Petrobras.
 
\section*{Data and code availability}
 The source code and data used in this work are available at \\ 
 \url{https://github.com/viluiz/gan}.

\bibliographystyle{unsrtnat} % orders according to appearance
\bibliography{references}

\end{document}